%% file: main.tex
\documentclass[11pt]{article}

\usepackage[preprint]{acl}

\usepackage{times}
\usepackage{latexsym}

\usepackage[T1]{fontenc}
\usepackage[utf8]{inputenc}

\usepackage{microtype}
\usepackage{inconsolata}
\usepackage{algorithm}
\usepackage{algpseudocode}
\usepackage{amsmath}
\usepackage{xcolor}
\usepackage{amssymb}
\usepackage{listings}
\usepackage{graphicx}
\usepackage{booktabs}
\usepackage{colortbl}
\usepackage{multirow}
\usepackage{makecell}

\definecolor{cforange}{RGB}{190,95,25}
\definecolor{cfrowgray}{gray}{0.94}
\title{CForce: Boosting Parallel Decoding for dLLMs via Consistency Forcing}

\author{
  Yuji Ren\textsuperscript{1,2*},\enspace
  Chenkai Xu\textsuperscript{1,2*},\enspace
  Zhuocheng Gong\textsuperscript{2},\enspace
  Jianguo Li\textsuperscript{2\textdagger},\enspace
  Zhijie Deng\textsuperscript{1\textdagger}
  \\
  \normalfont
  \textsuperscript{1}Shanghai Jiao Tong University
  \quad
  \textsuperscript{2}Ant Group
  \\
  \small
  \texttt{\{renyj26, 132435xck, zhijied\}@sjtu.edu.cn},\enspace
  \texttt{lijg.zero@antgroup.com}
}

\begin{document}
\maketitle
\begingroup
\renewcommand{\thefootnote}{\fnsymbol{footnote}}
\footnotetext[1]{Equal contribution. }
\footnotetext[2]{Corresponding authors.}
\endgroup

\input{sections/0}
\input{sections/1}
\input{sections/2}
\input{sections/preliminary}

\input{sections/method}

\input{sections/experiments}
\input{sections/conclusion}
\input{sections/limitations}
\input{sections/acknowledgments}

\bibliography{custom}

\clearpage
\appendix
\input{sections/apdx}

\end{document}

%% file: sections/0.tex
\begin{abstract}

Diffusion large language models (dLLMs) 
% enable parallel generation by revealing multiple masked tokens per forward pass, but aggressive threshold decoding can suffer from unreliable predictions at early low-context states. These states differ from the random or constructed mask patterns in existing training and distillation pipelines, so committing many tokens in parallel can lock in errors that later denoising steps would otherwise correct.
accelerate language generation by predicting multiple masks in a single forward pass.
However, existing dLLMs can suffer from unreliable predictions in early denoising stages under aggressive parallelism strategies, leading to errors that can propagate to later stages. 
To tackle this issue, 
% We propose \textbf{Consistency Forcing}, a post-training method that trains a dLLM on its own threshold-decoding trajectories. 
we present Consistency Forcing (CForce) for dLLMs, a distillation method to force the mask predictions of early stages to align with those of later stages. 
% By ``inference-aligned'', we mean that the model is trained with noise sampled from the model's on-policy denoising trajectories.
% We group self-generated states by accumulated mask-to-token reveals to form adjacent stage pairs, and force the output distribution at an earlier stage to match a stop-gradient prediction from the later, more informed stage. The objective combines confidence-adaptive KL alignment with token-level CE anchoring on student-masked positions, together with a curriculum strategy over stage difficulty. 
% This directly aligns the training states with the low-context states encountered at inference time and encourages staged consistency along the decoding trajectory. 
% Concretely, our approach explicitly promotes training-inference alignment by matching the low-context states encountered during generation.
% CForce follows an on-policy distillation formulation, where rollouts from the model itself are used as training inputs to improve training-inference alignment. 
CForce trains the model on pre-collected self-rollout trajectories, thereby improving training-inference alignment.
We introduce Confidence Adaptive KL Divergence as a distillation objective to conjoin the merits of forward and reverse KL.
% To improve training-inference alignment, we use on-policy  as training inputs, so the model is optimized on denoising trajectories that follow its own inference-time decoding behavior.
We further provide a theoretical analysis for the consistency objective to explain why CForce can approximately minimize the prediction error of early stages. % early prediction error to adjacent-stage distributional drift and reveal-
% boundary token error.
% This dramatically improves the effectiveness of the distillation.
Critically, the same formulation applies to both mask-to-token decoding and edit-capable decoding; in the edit-capable case, later token-to-token refinements provide additional supervision for earlier masked-state predictions. 
Experiments on non-edit and edit-capable LLaDA models show improved speed-quality trade-offs, especially under high-parallelism decoding budgets. 
Code is available at: \url{https://github.com/inclusionAI/dFactory}.
\end{abstract}

%% file: sections/1.tex
\section{Introduction}

\begin{figure}[t]
\centering
\includegraphics[width=\columnwidth]{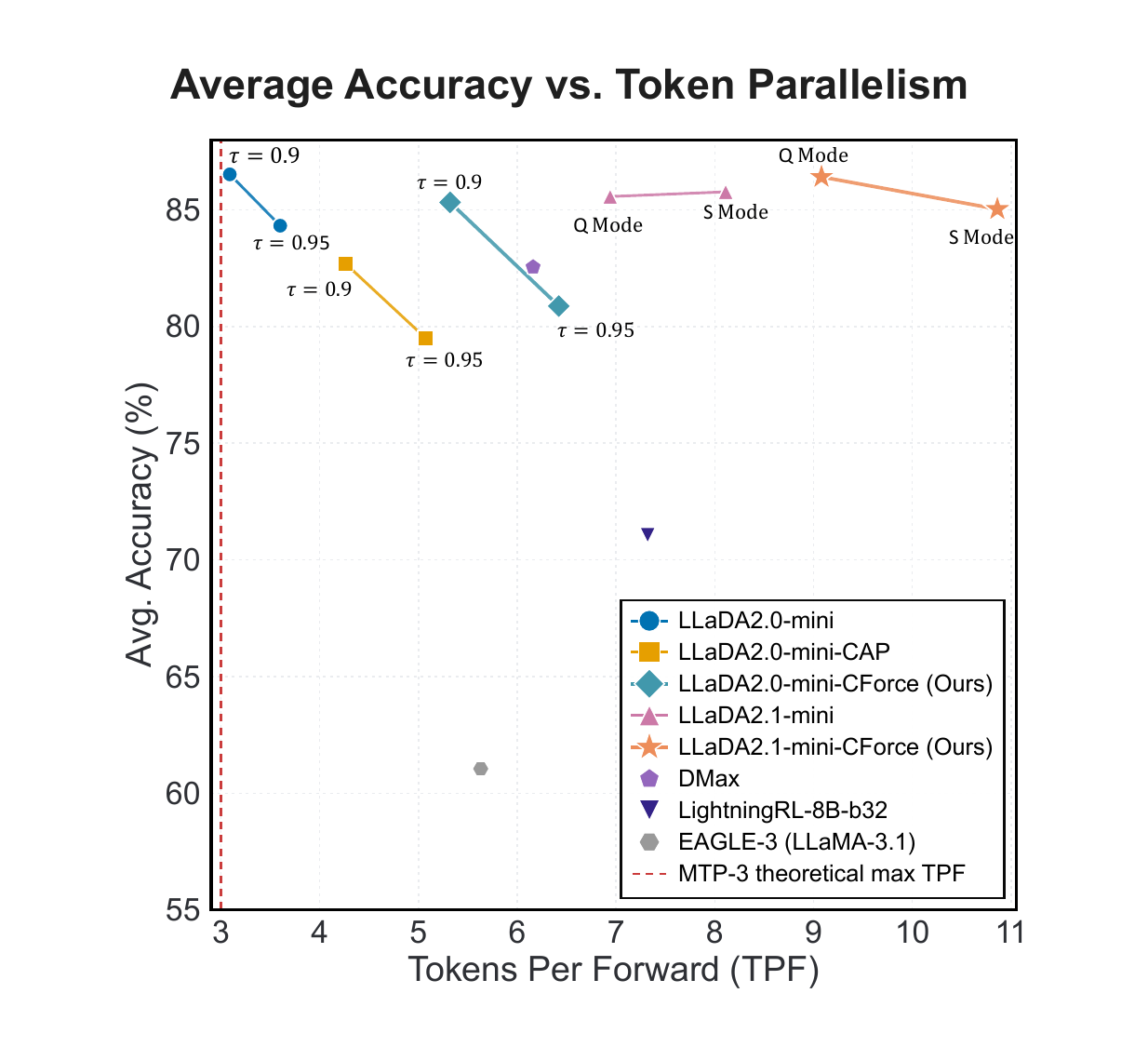}
\caption{\textbf{Speed--quality trade-off: accuracy vs.\ tokens per forward pass (TPF).} Upper-right is better. Consistency Forcing moves both models toward a higher-parallelism trade-off.}
\label{fig:main-effect}
\end{figure}

\begin{figure*}[t]
\centering
\includegraphics[width=\textwidth]{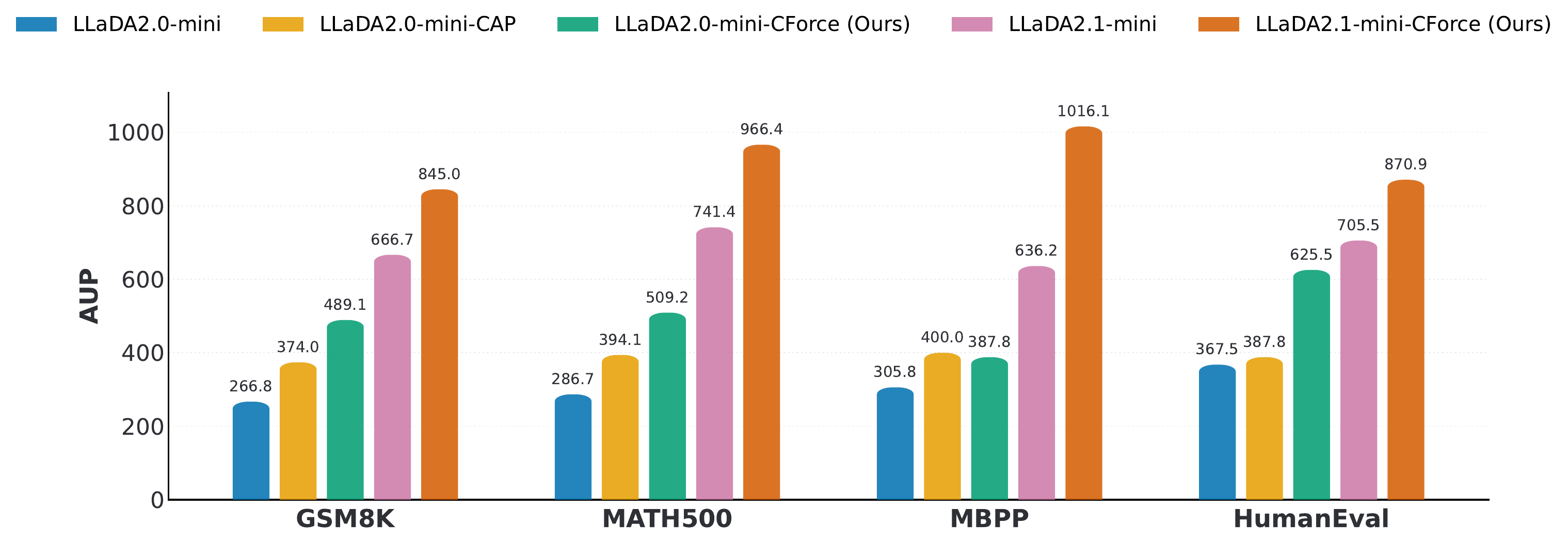}
\caption{\textbf{Per-task Accuracy Under Parallelism (AUP).} AUP integrates the score--parallelism curve into a single metric. CForce yields consistent gains across all four benchmarks.}
\label{fig:aup}
\end{figure*}

% Diffusion Large Language Models (dLLMs)~\citep{austin2023structured,zheng2024reparameterized,lou2024discrete,sahoo2024simple,nie2025large,ye2025dream7b,arriola2025block,gong2025diffucoder} offer a different route to fast text generation from autoregressive decoding. Instead of producing tokens strictly left to right, they iteratively denoise a partially masked sequence and can reveal many tokens in a single forward pass~\citep{wu2025fast,bie2025llada2,bie2026llada2,wang2025diffusion,cheng2025sdar}. 
% However, aggressive parallelism requires committing more tokens from early denoising stages, where limited context makes even high-confidence mask predictions unstable. Once such predictions are committed together, their errors can propagate to later stages and become difficult to recover. Thus, faster dLLM decoding depends not only on reducing denoising steps, but also on improving the reliability of early-stage predictions under high-parallelism decoding.

Diffusion large language models (dLLMs)~\citep{austin2023structured,zheng2024reparameterized,lou2024discrete,sahoo2024simple,nie2025large,ye2025dream7b,arriola2025block,gong2025diffucoder} generate text by iteratively denoising masked sequences, committing multiple tokens per forward pass to achieve high parallelism~\citep{wu2025fast,bie2025llada2,bie2026llada2,wang2025diffusion,cheng2025sdar}.
However, existing dLLMs can suffer from unreliable predictions in early denoising stages under aggressive parallelism strategies, and the resulting errors can propagate to later stages. 
Thus, faster dLLM decoding 
% should 
depend not only on reducing denoising steps, but also on improving the reliability of early-stage predictions under high-parallelism decoding.

This issue is especially important for edit-capable dLLMs. Conventional masked diffusion decoding is mainly mask-to-token (M2T), whereas edit-capable models such as LLaDA2.1~\citep{bie2026llada2} also perform token-to-token (T2T) refinement, allowing later denoising states to revise earlier drafts. These later states contain richer context and possible T2T corrections, which could provide stronger supervision for early M2T predictions before token commitment. Existing acceleration and distillation methods do not train this early-to-late consistency along the model's own threshold-decoding trajectory: some modify decoding or caching~\citep{kim2025klass, wu2025fast,liu2025dllmcache}, while others use constructed, teacher-generated, or privileged trajectories~\citep{zhang2026t3d,kim2025cdlm,liang2026cd4lm} whose information conditions differ from the student's inference states.

% We propose Consistency Forcing, a training method inspired by consistency models~\citep{salimans2022progressive,song2023consistency} that improves the speed-quality frontier of threshold-decoded dLLMs by aligning early low-context predictions with later higher-context stage on the model's own decoding path. 
We propose Consistency Forcing (CForce) for dLLMs, a distillation method inspired by consistency models~\citep{salimans2022progressive,song2023consistency} that forces the mask predictions of early stages to align with those of later stages. 
Specifically, 
starting from a pretrained dLLM, CForce trains the model on self-rollout trajectories, thereby improving training-inference alignment. 
% Then, CForce partitions each trajectory into stages, where each stage is formed after every \(S\) newly revealed masked tokens.
Then, CForce partitions each pre-collected trajectory into stages. Each stage boundary is set after a fixed number of newly revealed masked tokens, rather than at every native step, which would incur substantial trajectory-storage overhead.
For each adjacent pair, the earlier-stage prediction is trained to match a stop-gradient prediction from the same model at the later stage, without relying on a frozen teacher. 
Still-masked positions are aligned with \textbf{C}onfidence \textbf{A}daptive KL \textbf{D}ivergence (CAD), which dynamically interpolates between forward and reverse KL based on later-stage prediction confidence.
CForce also applies a cross-entropy (CE) anchor to stabilize token commitment.
A curriculum over reveal difficulty further stabilizes this adjacent-stage objective~\citep{unicm2026,liu2024ccm}.
% We collect threshold-decoding trajectories from a pretrained dLLM, divide them into adjacent stages, and train each earlier stage to match a stop-gradient copy of the same model at the next, more informed stage, without relying on a frozen teacher model. 
% Still-masked positions are aligned with Confidence-adaptive KL Divergence (CAD) to control distributional drift, while newly revealed positions receive a cross-entropy anchor to stabilize token commitment. A curriculum over reveal difficulty further stabilizes this adjacent-stage objective~\citep{unicm2026,liu2024ccm}.

% We further provide a theoretical analysis showing that early-stage prediction error can be upper-bounded by adjacent-stage drift and reveal-boundary error, directly matching the two components of our objective (Section~\ref{sec:theory}).
We further provide a theoretical analysis supporting this design: early-stage prediction error is upper-bounded by adjacent-stage distributional drift plus reveal-boundary token error (Section~\ref{sec:theory}), two terms that correspond directly to the CAD and CE components of our objective.

Empirically, the clearest result appears on the edit-capable LLaDA2.1-mini~\citep{bie2026llada2}: Consistency Forcing increases average Tokens Per Forward (TPF) from 6.94 to 9.08 while improving average accuracy from 85.57 to 86.41. On the non-edit LLaDA2.0-mini~\citep{bie2025llada2}, the method shows an explicit speed-quality trade-off, increasing average TPF from 3.60 to 6.42 and improving few-step accuracy at fixed TPF budgets under aggressive decoding. Figure~\ref{fig:main-effect} and Figure~\ref{fig:aup} visualize this effect.

Our contributions are fourfold: (1) we reframe dLLM acceleration around early-stage reliability under aggressive threshold decoding, rather than focusing only on fewer sampling steps; (2) we introduce Consistency Forcing, which aligns adjacent stages on the model's own decoding trajectory using CAD and a CE anchor; (3) we instantiate this idea in edit-capable dLLMs, where later T2T-refined stages supervise earlier M2T predictions; and (4) we show that the same recipe improves few-step and high TPF behavior on a non-edit dLLM, with an explicit speed-quality trade-off.

%% file: sections/2.tex
\section{Related Work}
\label{sec:related}

\paragraph{Masked Diffusion Language Models.}
Masked diffusion language models (MDLMs) cast text generation as discrete denoising~\citep{austin2023structured,zheng2024reparameterized,lou2024discrete}. Building on earlier masked generation methods~\citep{ghazvininejad2019maskpredict,chang2022maskgit}, recent dLLMs further scale this paradigm, enabling stronger text and code generation~\citep{nie2025large,ye2025dream7b,gong2025diffucoder}. To improve generation speed, several dLLMs introduce confidence-based parallel decoding~\citep{wu2025fast,wang2025diffusion,cheng2025sdar,bie2025llada2,chen2026dmax}, while edit-capable models allow later denoising states to revise committed tokens~\citep{bie2026llada2}. These later corrections are especially relevant under aggressive parallel decoding, where early M2T predictions are made with limited context and can benefit from supervision induced by later denoising states.

\paragraph{Acceleration Methods for dLLMs.}
Existing acceleration methods follow several complementary directions. Inference-time approaches improve token selection or verification without changing model parameters~\citep{kim2025klass,xu2025lopa,agrawal2025spiffy,gao2025self}, while system-level approaches reduce computation through caching~\citep{wu2025fast,ma2025dkvcache,liu2025dllmcache}. Training-based approaches instead improve few-step generation with constructed, teacher-generated, or privileged trajectories~\citep{dparallel2025,zhang2026t3d,kim2025cdlm,liang2026cd4lm}. These methods accelerate decoding but do not directly train the model on the low-context states produced by its own threshold-decoding path.

\paragraph{Consistency Distillation in Diffusion Models.}
Diffusion distillation reduces sampling cost by compressing slow multi-step generation into faster samplers or models~\citep{luhman2021knowledge,salimans2022progressive,meng2022distillation}. Consistency models extend this idea by enforcing agreement across trajectory states, enabling one-step or few-step generation~\citep{song2023consistency,luo2023latent,kou2024cllm}. Related objectives have also been studied for discrete generative models~\citep{hayakawa2025di4c,sahoo2025duo,unicm2026} and, more recently, for dLLMs~\citep{zhang2026t3d,kim2025cdlm,liang2026cd4lm}. These works demonstrate the usefulness of cross-state supervision. However, they typically depend on separate teachers, fixed representations, or teacher-generated targets. In contrast, Consistency Forcing constructs adjacent-stage constraints from the model's own threshold-decoding trajectory without a frozen teacher, and adapts them to both masked-token reveal and edit transitions.

% \paragraph{On-policy distillation.}
% Conventional language-model distillation uses teacher soft targets or generated sequences~\citep{hinton2015distilling,kim2016sequence}. On-policy distillation instead applies teacher feedback on student-generated trajectories to reduce train-inference mismatch~\citep{agarwal2024policy,gu2023minillm,li2026rethinking}. A useful feature of these methods is reverse KL, which encourages mode-seeking alignment with high-probability teacher tokens~\citep{agarwal2024policy,gu2023minillm,li2026rethinking}. This motivates us to go beyond a traditional single forward-KL objective: we introduce a confidence-adaptive KL that strengthens reverse KL on confident later-state predictions, while keeping forward KL dominant for uncertain targets, following recent uncertainty-aware and hybrid distillation ideas~\citep{jin2026entropy,zhu2026hybrid}.

%% file: sections/preliminary.tex
\section{Preliminaries}
\label{sec:preliminaries}

% \begin{figure}[t]
% \centering
% \includegraphics[width=\columnwidth]{figs/fig2.pdf}
% \caption{\textbf{Comparison between Threshold Decoding and Edit-capable Decoding.} Edit-capable decoding provides later T2T correction signals that can guide earlier M2T predictions.}
% \label{fig:decoding-methods}
% \end{figure}
\begin{figure}[t]
\centering
\includegraphics[width=\columnwidth]{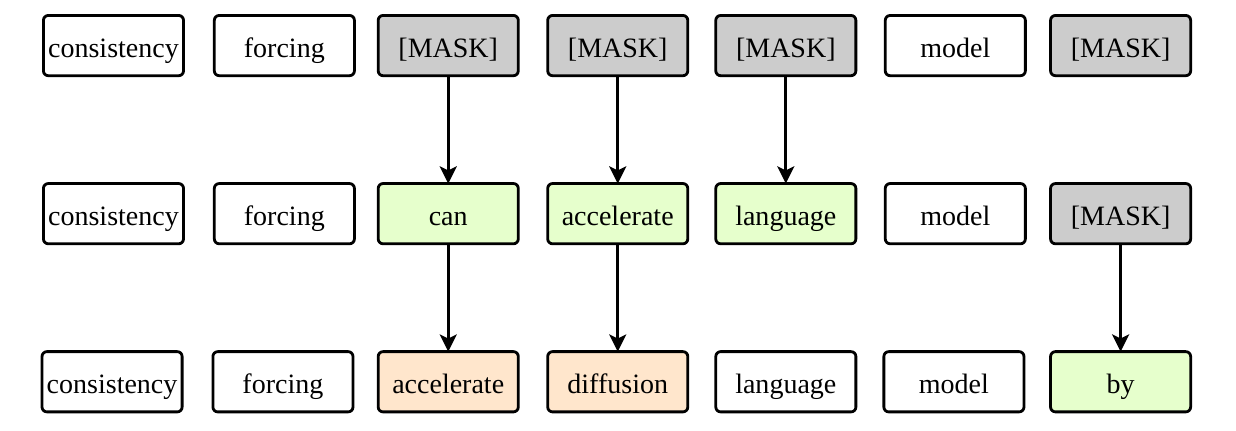}
\caption{An illustration of edit-capable dLLM decoding. At each decoding step, high-confidence masked positions are denoised, while already generated tokens may also be revised to correct earlier mistakes.}
\label{fig:decoding-methods}
\end{figure}

\paragraph{Masked Diffusion Language Models.}

\begin{figure*}[t]
\centering
\includegraphics[width=\textwidth]{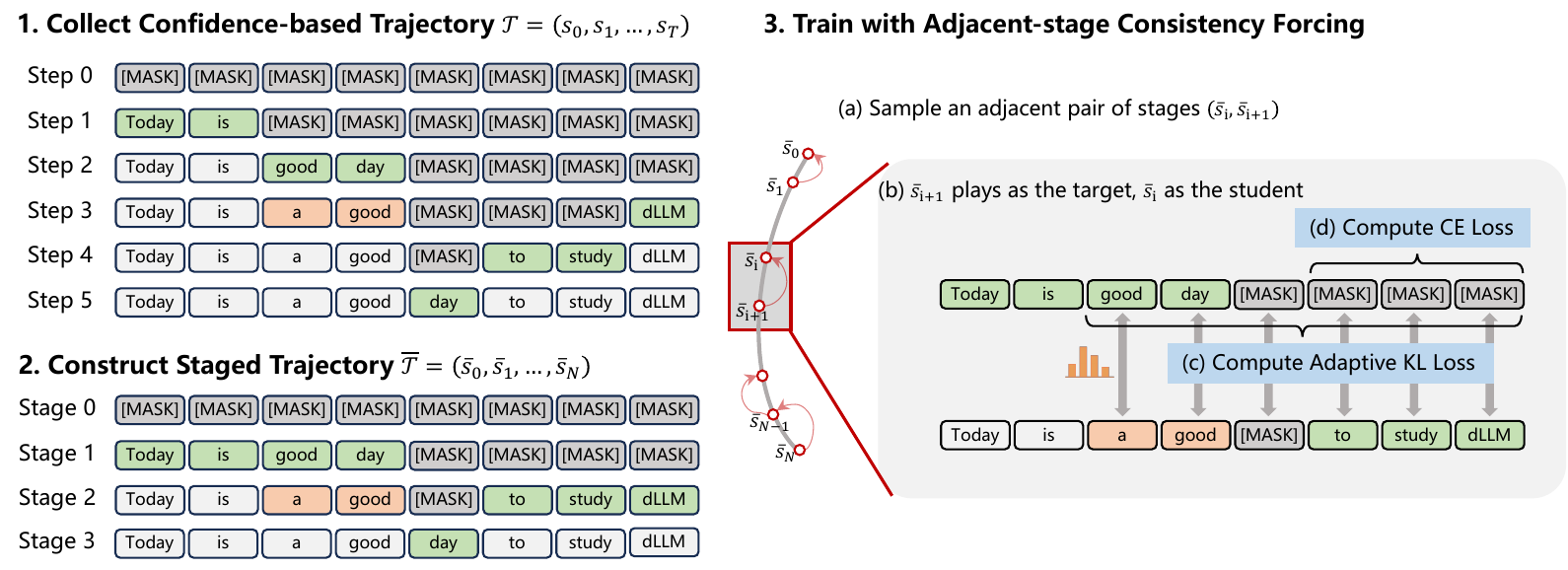}
\caption{\textbf{Overview of Consistency Forcing.} Left: self-rollout threshold-decoding trajectories are downsampled into adjacent stage pairs whenever the cumulative number of M2T reveals reaches the stage size $S$; right: the earlier low-context state is trained against same-model stop-gradient predictions from the later, more informative state.}
\label{fig:method-overview}
\end{figure*}

Masked diffusion language models formulate text generation as denoising in a discrete token space~\citep{austin2023structured,lou2024discrete,sahoo2024simple}. Given a prompt $x\in\mathcal{X}$ and a clean target sequence $y=(y^1,\ldots,y^L)$ over vocabulary $\mathcal{V}$, the forward process corrupts $y$ by independently replacing tokens with $[\mathrm{MASK}]$. Let $\alpha_t\in[0,1]$ denote the probability that a token remains unmasked at noise time $t$, and let $s_t$ be the corrupted sequence. Under the absorbing-mask process,
\[
q(s_t^j \mid y^j)=
\begin{cases}
\alpha_t, & s_t^j = y^j,\\
1-\alpha_t, & s_t^j = [\mathrm{MASK}].
\end{cases}
\]

The reverse model predicts a categorical distribution
$q_\theta(\cdot\mid x,s_t,j)$ for each masked position $j$.

Let $M_t=\{j:s_t^j=[\mathrm{MASK}]\}$ be the masked positions.
Training can be written as the standard variational bound for masked diffusion, which reduces to a weighted cross-entropy over these positions:
\begin{equation}\label{eq:mdlm}
\begin{aligned}
\mathcal{L}_{\mathrm{MDLM}}
&=
\mathbb{E}_{(x,y),t,s_t}
\left[
w(t)\sum_{j\in M_t}
\mathrm{CE}_{\theta}^j
\right],
\\
\mathrm{CE}_{\theta}^j
&=
-\log q_\theta(y^j\mid x,s_t,j).
\end{aligned}
\end{equation}
Here $w(t)=-\alpha'_t/(1-\alpha_t)$ is the positive ELBO weight induced by the absorbing diffusion process. This objective trains the model to recover clean tokens from partially observed contexts, enabling multiple positions to be denoised in parallel at inference time.

In current large-scale dLLMs, this masked diffusion objective is commonly used together with block diffusion~\citep{arriola2025block,bie2025llada2}, which partitions the target sequence into blocks and denoises each block in parallel while generating blocks sequentially. 
% In the rest of this paper, we follow this block-diffusion setting and discuss decoding and training within a block unless otherwise specified.
Recent edit-capable dLLMs further augment this block-diffusion framework with token-to-token refinement, allowing committed tokens to be revised in later denoising states~\citep{bie2026llada2}. 
% In the rest of this paper, 
Hereafter, 
we follow this block-diffusion setting and discuss decoding and training within a block unless otherwise specified.

% MDLMs formulate text generation as a discrete denoising process~\citep{austin2023structured,lou2024discrete,sahoo2024simple}. Given a prompt $x \in \mathcal{X}$ and a clean target sequence $y=(y^1,\ldots,y^L)$ over vocabulary $\mathcal{V}$, the forward process samples a noise level $\alpha$ and replaces a random subset of target tokens with $[\mathrm{MASK}]$, producing a corrupted state $s_\alpha$. The reverse model predicts a categorical distribution $q_\theta(\cdot \mid x,s_\alpha,j)$ for each masked position. Following the standard masked diffusion objective, training minimizes a weighted masked-token reconstruction loss:
% \begin{equation}\label{eq:mdlm}
% \small
% \begin{aligned}
% \mathcal{L}_{\mathrm{MDLM}}
% =
% \mathbb{E}_{y,\alpha,s_\alpha}
% \Bigg[
% -w(\alpha)
% \sum_{j\in M_\alpha}
% \log q_\theta(y^j \mid x,s_\alpha,j)
% \Bigg],
% \end{aligned}
% \end{equation}
% where $M_\alpha=\{j:s_\alpha^j=[\mathrm{MASK}]\}$ and $w(\alpha)$ denotes the noise-level weight. This objective trains the model to recover clean tokens from partially masked contexts, providing the basis for parallel denoising at inference time.

\paragraph{Confidence-based Parallel Decoding.}
At inference time, block diffusion generates one block by starting from an all-mask block state $s_0$ and iteratively denoising it into the all-clean state. This process produces a trajectory $\mathcal{T}=(s_0,s_1,\ldots,s_T)$, where $T$ is the number of decoding steps allocated to the block. At step $t$, let
$M_t=\{j:s_t^j=[\mathrm{MASK}]\}$ denote the positions that remain masked, and let
$p_t^j(v)=q_\theta(v\mid x,s_t,j)$ be the model distribution at position $j\in M_t$.

Modern dLLMs reveal multiple positions in each forward pass according to model confidence~\citep{bie2025llada2}. The confidence of a masked position is defined as $c_t^j=\max_{v\in\mathcal{V}}p_t^j(v)$. Given a threshold $\tau$, all positions whose confidence exceeds the threshold are selected for commitment. To ensure progress, if no position exceeds the threshold, the most confident remaining position is selected:
\begin{equation}\label{eq:threshold}
\small
\Delta M_t =
\begin{cases}
\{j\in M_t:c_t^j>\tau\},
& \text{if } \{j\in M_t:c_t^j>\tau\}\neq\emptyset,\\
\{\arg\max_{j\in M_t} c_t^j\},
& \text{otherwise}.
\end{cases}
\end{equation}
For each selected position $j\in\Delta M_t$, the decoder commits the most probable token,
\begin{equation}
s_{t+1}^j=\arg\max_{v\in\mathcal{V}}p_t^j(v),
\end{equation}
and the remaining positions stay masked. Equivalently,
\begin{equation}
M_{t+1}=M_t\setminus \Delta M_t .
\end{equation}

% \paragraph{Confidence-threshold Decoding.}
% At inference time, an MDLM starts from an all-mask state $s_0$ and iteratively produces a denoising trajectory $\tau=(s_0,s_1,\ldots,s_T)$ until reaching the final text $s_T$. At decoding step $t$, let $M_t=\{j:s_t^j=[\mathrm{MASK}]\}$ and $p_t^j(v)=q_\theta(v\mid x,s_t,j)$. To accelerate generation, modern dLLMs employ confidence-aware parallel decoding~\citep{bie2025llada2}. The confidence of each masked position is $c_t^j=\max_{v\in\mathcal{V}}p_t^j(v)$. A position is committed to its most probable token if its confidence exceeds threshold $\tau$. To guarantee generation progress, the newly revealed positions $\Delta M_t$ are defined as:
% \begin{equation}\label{eq:threshold}
% \small
% \Delta M_t = 
% \begin{cases} 
% \{j \in M_t : c_t^j > \tau\}, & \text{if } \{j \in M_t : c_t^j > \tau\} \neq \emptyset \\ 
% \{ \arg\max_{j \in M_t} c_t^j \}, & \text{otherwise}
% \end{cases}
% \end{equation}
% The masked set is then updated as:
% \begin{equation}
% M_{t+1} = M_t \setminus \Delta M_t.
% \end{equation}
% This threshold-based mechanism progressively accumulates rich context for the remaining masks, fundamentally distinguishing it from sequential autoregressive decoding.

\paragraph{Edit-capable Diffusion Decoding.}
Recent edit-capable dLLMs extend the confidence-based decoding process by allowing T2T refinement after M2T commitment~\citep{bie2026llada2}. Given an edit threshold $\tau_{\mathrm{edit}}$, positions selected for editing are
\begin{equation}\label{eq:edit}
\small
E_t=
\left\{
j\notin M_t:
\max_{v\in\mathcal{V},\,v\neq s_t^j}
q_\theta(v\mid x,s_t,j)
>
\tau_{\mathrm{edit}}
\right\}.
\end{equation}
For each $j\in E_t$, the decoder replaces the current token with the most probable alternative,
\begin{equation}
s_{t+1}^j
=
\arg\max_{v\in\mathcal{V},\,v\neq s_t^j}
q_\theta(v\mid x,s_t,j).
\end{equation}

As illustrated in Figure~\ref{fig:decoding-methods}, in edit-capable decoding, later states can correct tokens that were committed earlier. These T2T refinements are especially important for fast decoding when the M2T threshold $\tau$ is low. Consistency Forcing uses this later-stage refinement signal to supervise earlier masked-state predictions.

% \paragraph{Edit-capable Diffusion Decoding.}
% In edit-capable dLLMs such as LLaDA2.1~\citep{bie2026llada2}, this paradigm extends beyond monotonic M2T generation to support a draft-and-edit process. This capability is natively acquired during training by optimizing a mixture of M2T and T2T objectives. During decoding, the model re-evaluates previously committed tokens using a dedicated editing threshold $\tau_{\mathrm{edit}}$. For a committed position $j \notin M_t$, if an alternative token different from its current token in $s_t$ achieves a probability exceeding $\tau_{\mathrm{edit}}$, the position is selected for revision. Formally, the set of positions designated for editing is defined as:
% \begin{equation}\label{eq:edit-set}
% \small
% E_t =
% \{j \notin M_t : \max_{v \neq s_t^j} p_t^j(v) > \tau_{\mathrm{edit}}\}.
% \end{equation}

% As illustrated in Figure~\ref{fig:decoding-methods}, this T2T reasoning capability lays the core mechanistic foundation for our subsequent application of Consistency Forcing, where reliable later-stage predictions are utilized to constrain and guide early-stage M2T predictions.

%% file: sections/method.tex
\section{Method}
\label{sec:method}

% --- Old content ---
% Consistency Forcing trains a dLLM to make reliable predictions from earlier, less informative denoising states. The method targets the reliability bottleneck discussed in the introduction: under aggressive threshold decoding, many tokens must be committed before much surrounding context has been revealed. We construct supervision from cached self-generated threshold-decoding transitions produced by the pretrained initialization. For each adjacent pair, the model at the earlier state is trained to match a stop-gradient copy of the same model evaluated at the later, more informative state.
% --- End of old content ---

\subsection{Overview}
Consistency Forcing trains a dLLM to make more reliable predictions from early, low-context denoising states. The method collects threshold-decoding trajectories from the pretrained initialization $q_{\theta_0}$, converts them into staged adjacent pairs, and uses the later stage as a same-model stop-gradient target for the earlier stage. The training objective combines adjacent-stage distributional alignment, a reveal-boundary CE anchor, and a curriculum over transition difficulty; building on this, edit-capable models can further feed later T2T signals back into earlier states to improve M2T drafting. Figure~\ref{fig:method-overview} illustrates the overall pipeline.

% To resolve the reliability bottleneck and the lack of early-stage supervision mentioned above, we propose Consistency Forcing. Inspired by consistency models~\citep{song2023consistency}, it aligns early low-context predictions with later higher-context predictions along the model's own threshold-decoding trajectory. However, supervising every decoding step is often ineffective, since a single threshold-decoding step may reveal only a few tokens and thus provides little contextual change. We therefore form coarser adjacent stages by grouping consecutive decoding steps. These stage pairs provide a meaningful context gap for consistency supervision, and are constructed from self-generated decoding trajectories without introducing external data or a separately trained teacher. Figure~\ref{fig:method-overview} summarizes the Consistency Forcing pipeline.

\subsection{Staged Trajectory Construction}
\label{sec:trajectory}

We need to construct trajectories used for consistency forcing.
% We construct trajectories using the confidence-based decoding process described in Section~\ref{sec:preliminaries}. 
Specifically, we first collect the original confidence-based decoding trajectory $\mathcal{T}=(s_0,s_1,\ldots,s_T)$ using the cold-start initialization $q_{\theta_0}$ as described in Section~\ref{sec:preliminaries}. Rather than treating every consecutive native transition in $\mathcal{T}$ as a training pair, we use the native rollout only as the source trajectory: dense native-step supervision is inefficient and often provides only a weak training signal, because adjacent native states may differ by very few newly revealed tokens.

Next, we maintain a reveal counter along the rollout. Whenever the cumulative number of newly revealed tokens reaches a fixed stage size $S$, we save the current state as a stage boundary and reset the counter. The initial all-mask state is always saved, and the final decoded state is appended if it has not already been saved. This produces a staged trajectory
$\bar{\mathcal{T}}=(\bar{s}_0,\bar{s}_1,\ldots,\bar{s}_N)$,
where adjacent stages are separated by cumulative M2T progress rather than by the raw native update index.

We then form adjacent pairs from $\bar{\mathcal{T}}$. Downsampling by cumulative M2T reveals forms coarser stages, so each adjacent pair contains a more substantial change in visible context.

For each adjacent stage pair $(\bar{s}_i,\bar{s}_{i+1})$, let
$\bar{M}_i=\{j:\bar{s}_i^j=[\mathrm{MASK}]\}$ denote the positions masked in the earlier stage. We split these positions into
\begin{equation}
\mathcal{U}_i = \bar{M}_i \cap \bar{M}_{i+1},
\qquad
\Delta_i = \bar{M}_i \setminus \bar{M}_{i+1}.
\end{equation}
Here, $\mathcal{U}_i$ contains positions that remain masked in both stages, while $\Delta_i$ contains positions that are revealed between the two stages. These sets define the supports of the CAD and CE terms below.

For edit-capable decoding, we also record native edit events during the trajectory rollout and aggregate them between adjacent stage boundaries; the resulting stage-level edit set $\bar{E}_i$ is introduced in Section~\ref{sec:edit}. Since all stages are obtained from the model's own confidence-based decoding process, the staged trajectories preserve the mask, reveal, and edit patterns that arise during fast inference.

% and downsample native decoding steps by cumulative M2T reveals. Whenever the reveal counter reaches a fixed stage size $S$, the current state is saved as a stage boundary. The saved boundaries define a staged training trajectory $\bar{\mathcal{T}} = (\bar{s}_0, \bar{s}_1, \ldots, \bar{s}_N)$, from which we form adjacent-stage transitions.

% For any adjacent pair of stages $(\bar{s}_i, \bar{s}_{i+1})$, let $\bar{M}_i = \{j : \bar{s}_i^j = [\mathrm{MASK}]\}$ denote the masked positions at stage $i$. We partition the positions into two disjoint groups:
% \begin{equation}
% \mathcal{U}_i = \bar{M}_i \cap \bar{M}_{i+1}, \qquad
% \Delta_i = \bar{M}_i \setminus \bar{M}_{i+1}.
% \end{equation}
% Here, $\mathcal{U}_i$ contains positions that remain masked in both stages, while $\Delta_i$ is the stage-level newly revealed set. The complete construction algorithm is provided in Appendix~\ref{app:trajectory-algorithm}; for edit-capable decoding, native edit sets $E_t$ are also recorded and later aggregated into $\bar{E}_i$ in Section~\ref{sec:edit}. Since stages are sampled from threshold-decoding trajectories, the resulting pairs retain the mask and edit patterns encountered at inference time.

\subsection{Adjacent-stage Consistency Forcing}
\label{sec:forcing}

For each sampled adjacent stage pair $(\bar{s}_i, \bar{s}_{i+1})$, we use the earlier stage as the student input and the later, more informative stage as the target context. The target branch is evaluated under stop-gradient, denoted by $\mathrm{sg}[\cdot]$, so the model learns to approximate from $\bar{s}_i$ the prediction it would make after receiving the additional context in $\bar{s}_{i+1}$. We instantiate this adjacent-stage constraint with two complementary terms: CAD controls soft distributional drift on positions that remain latent across the transition, while a CE anchor stabilizes positions at the M2T reveal boundary. This decomposition follows the error-bound motivation in Section~\ref{sec:theory}.

\paragraph{Confidence Adaptive KL Divergence.}
On $\mathcal{U}_i$, both stages are still masked, so we align the student distribution $p_{\mathrm{stu}}^j=q_\theta(\cdot\mid x,\bar{s}_i,j)$ with the stop-gradient target distribution $p_{\mathrm{tar}}^j=\mathrm{sg}[q_\theta(\cdot\mid x,\bar{s}_{i+1},j)]$. A standard forward KL minimizes drift but can be weak at sharpening predictions. Conversely, reverse KL is a strong mode-seeking objective but risks premature collapse when applied to uncertain targets. 
To combine their strengths, we design Confidence Adaptive KL Divergence (CAD), which dynamically mixes forward and reverse KL based on the later stage's prediction confidence $c_j=\max_v p_{\mathrm{tar}}^j(v)$, defined as
\begin{equation}\label{eq:cad}
\small
\mathcal{L}_{\mathrm{CAD}}
= \frac{1}{|\mathcal{U}_i|}
\sum_{j \in \mathcal{U}_i}
\left[
D_{\mathrm{KL}}(p_{\mathrm{tar}}^j \| p_{\mathrm{stu}}^j)
+ c_j D_{\mathrm{KL}}(p_{\mathrm{stu}}^j \| p_{\mathrm{tar}}^j)
\right],
\end{equation}
where the forward KL term stably controls distributional drift connected to the upper bound in Section~\ref{sec:theory}, while the confidence-weighted reverse KL adds mode-seeking pressure only when the later-stage prediction is confident.

\paragraph{CE Anchor.}
On the newly revealed positions $\Delta_i$, we apply a cross-entropy anchor to stabilize token commitments:
\begin{equation}\label{eq:ce}
\mathcal{L}_{\mathrm{CE}}
= -\frac{1}{|\Delta_i|}
\sum_{j \in \Delta_i}
\log q_\theta(y^j \mid x, \bar{s}_i, j),
\end{equation}
where \(y^j\) denotes the clean target token at position \(j\). By anchoring the student prediction immediately before reveal, the CE term discourages errors at the boundary where masked positions become committed tokens.

\paragraph{Edit-capable dLLMs.}
\label{sec:edit}
As described in Section~\ref{sec:preliminaries}, edit-capable dLLMs provide an additional source of later-stage supervision: T2T edits of already visible tokens. We define the stage-level edit set $\bar{E}_i$ as the positions whose visible tokens differ between adjacent stages $\bar{s}_i$ and $\bar{s}_{i+1}$:
\begin{equation}\label{eq:stage-edit}
\small
\begin{aligned}
\bar{E}_i =
\{j :\;& \bar{s}_i^j \neq [\mathrm{MASK}],\;
\bar{s}_{i+1}^j \neq [\mathrm{MASK}], 
& \bar{s}_i^j \neq \bar{s}_{i+1}^j\}.
\end{aligned}
\end{equation}
Edit-capable trajectories contain both M2T reveals and later T2T corrections, so we broaden the CAD domain from $\mathcal{U}_i$ to $\bar{M}_i \cup \bar{E}_i$. The $\bar{M}_i$ term aligns student-masked positions with later soft predictions under additional revealed context, 
while $\bar{E}_i$ lets visible tokens that change between stages contribute T2T correction signals. Since these later-stage soft distributions provide broader supervision, we also extend the CE anchor from $\Delta_i$ to all student-masked positions $\bar{M}_i$. This uses edit-capable trajectories without changing the decoding algorithm.

\paragraph{Curriculum Transition.}
\label{sec:curriculum}
At the beginning of optimization, the full stage gap can be difficult because the student has not yet learned to predict tokens that are revealed many positions ahead. We therefore expose the later-stage context gradually. At optimization step $u$, only a fraction $\rho_u$ of the newly revealed positions in $\Delta_i$ are used to form the later-stage context:
\begin{equation}\label{eq:curriculum}
\rho_u = \min\!\left(1,\; \rho_0 + (1 - \rho_0)\frac{u}{T_{\mathrm{train}}}\right),
\end{equation}
where $\rho_0=0.1$ and $T_{\mathrm{train}}$ is the total training steps. Concretely, we select a subset $\Gamma_i(u)\subseteq\Delta_i$ with $|\Gamma_i(u)|=\lfloor\rho_u|\Delta_i|\rfloor$ and form a partially revealed later state by revealing only $\Gamma_i(u)$ while keeping the remaining positions in $\Delta_i$ masked. Initially, the target state is close to the student state, stabilizing local consistency learning; as $\rho_u$ increases, the model is gradually exposed to larger adjacent-stage transitions and eventually learns the consistency constraint used by fast decoding.

\paragraph{Overall Objective.}
The training loss is
\begin{equation}\label{eq:total-loss}
\mathcal{L}
= \mathcal{L}_{\mathrm{CAD}}
+ \lambda_{\mathrm{CE}}\mathcal{L}_{\mathrm{CE}},
\end{equation}
where $\lambda_{\mathrm{CE}}$ balances distributional forcing and token anchoring.

\subsection{Theoretical Analysis}
\label{sec:theory}

\paragraph{Setup.}
We keep the main text focused on the intuition and key conclusion, and defer the full derivation to Appendix~\ref{app:theory}. The goal is to justify why the Consistency Forcing objective targets early-stage reliability. For two categorical distributions $p$ and $q$ over vocabulary $\mathcal{V}$, the total variation (TV) distance is defined as
\begin{equation}\label{eq:tv-distance}
D_{\mathrm{TV}}(p,q)
= \frac{1}{2}\sum_{v\in\mathcal{V}} |p(v)-q(v)|.
\end{equation}
Consider a position $j$ revealed at stage $r_j$, with clean target token $y^j=\bar{s}_N^j$, and let $p_i^j=q_\theta(\cdot\mid x,\bar{s}_i,j)$. For any earlier masked stage $i<r_j$, the prediction error at $\bar{s}_i$ is measured as $D_{\mathrm{TV}}(p_i^j,\delta_{y^j})$, where $\delta_{y^j}$ denotes the one-hot distribution on $y^j$.

\paragraph{Main Bound.}
The early-stage prediction error can be bounded by adjacent-stage distributional drift and the reveal-boundary token error:
\begin{equation}\label{eq:main-bound}
\begin{split}
D_{\mathrm{TV}}(p_i^j,\delta_{y^j})
&\le
\sum_{r=i}^{r_j-2}
\sqrt{\frac{1}{2}D_{\mathrm{KL}}(p_{r+1}^j\|p_r^j)} \\
&\quad - \log p_{r_j-1}^j(y^j).
\end{split}
\end{equation}
The first term measures how much the model prediction drifts as more context is revealed before position $j$ is committed, while the second term measures the token-level error immediately before reveal, which is controlled by the CE loss on the clean token under commitment.

\paragraph{Connection to CForce.}
Eq.~\eqref{eq:main-bound} directly matches our objective: the forward KL component in CAD reduces adjacent-stage drift, while the CE anchor controls the reveal-boundary error. The confidence-weighted reverse KL in CAD is non-negative and therefore preserves this forward KL alignment while sharpening predictions when the later-stage target is confident. Thus, Consistency Forcing optimizes a tractable surrogate for the upper bound on early-stage prediction error.

%% file: sections/experiments.tex
\begin{table*}[t]
\centering
\scriptsize
\setlength{\tabcolsep}{2.2pt}
\resizebox{\textwidth}{!}{%
% \small{
\begin{tabular}{@{}lll*{4}{ccc}@{}}
\toprule
\multirow{2}{*}{\textbf{Diffusion type}} & \multirow{2}{*}{\textbf{Edit?}} & \multirow{2}{*}{\textbf{Model}}
& \multicolumn{3}{c}{\textbf{GSM8K}}
& \multicolumn{3}{c}{\textbf{MATH500}}
& \multicolumn{3}{c}{\textbf{MBPP}}
& \multicolumn{3}{c}{\textbf{HumanEval}} \\
\cmidrule(lr){4-6}\cmidrule(lr){7-9}\cmidrule(lr){10-12}\cmidrule(l){13-15}
& & & \textbf{Score} & \textbf{TPF} & \textbf{AUP}
& \textbf{Score} & \textbf{TPF} & \textbf{AUP}
& \textbf{Score} & \textbf{TPF} & \textbf{AUP}
& \textbf{Score} & \textbf{TPF} & \textbf{AUP} \\
\midrule
\multirow{4}{*}{\makecell[l]{Pure\\diffusion}} & \multirow{4}{*}{$\times$}
 & dUltra-coding-b32 & 81.52\textsuperscript{*} & 8.40 & -- & 35.64 & 6.72\textsuperscript{*} & -- & 37.04 & 7.29\textsuperscript{*} & -- & 35.85 & 7.23\textsuperscript{*} & -- \\
 & & d3LLM-LLaDA & 73.09 & 9.11\textsuperscript{*} & 637.65\textsuperscript{*} & 30.36 & 5.74 & 107.64\textsuperscript{*} & 40.60 & 4.21 & 88.36 & 39.63 & 5.95 & 96.64 \\
 & & d3LLM-Dream & 81.36 & 4.94 & 391.33 & 38.21\textsuperscript{*} & 3.92 & 97.50 & 55.60\textsuperscript{*} & 2.96 & 141.41\textsuperscript{*} & 57.10\textsuperscript{*} & 3.20 & 129.48\textsuperscript{*} \\
 & & D2F-LLaDA & 74.39 & 2.88 & 213.76 & 28.94 & 2.66 & 49.00 & 39.00 & 2.13 & 52.96 & 40.64 & 2.69 & 61.98 \\
\midrule
\multirow{7}{*}{\makecell[l]{Block\\diffusion}} & \multirow{4}{*}{$\times$}
 & LightningRL-8B-b32 & 90.30 & \textbf{5.58} & \textbf{492.40} & 63.00 & \underline{6.28} & \underline{407.50} & 58.30 & \textbf{11.10} & \textbf{641.60} & 72.60 & \underline{6.30} & \underline{450.10} \\
 & & LLaDA2.0-mini & \textbf{93.25} & 2.85 & 266.82 & \textbf{81.80} & 3.46 & 286.71 & \textbf{78.69} & 3.80 & 305.84 & \textbf{83.54} & 4.28 & 367.52 \\
 & & LLaDA2.0-mini-CAP & \underline{91.74} & 4.09 & 374.03 & \underline{81.00} & 4.85 & 394.10 & 72.13 & 5.43 & \underline{400.00} & 73.18 & 5.89 & 387.79 \\
 & & \cellcolor{cfrowgray}LLaDA2.0-mini-CForce (Ours) & \cellcolor{cfrowgray}\underline{91.74} & \cellcolor{cfrowgray}\underline{5.36} & \cellcolor{cfrowgray}\underline{489.07} & \cellcolor{cfrowgray}79.20 & \cellcolor{cfrowgray}\textbf{6.35} & \cellcolor{cfrowgray}\textbf{509.23} & \cellcolor{cfrowgray}\underline{73.30} & \cellcolor{cfrowgray}\underline{6.00} & \cellcolor{cfrowgray}387.80 & \cellcolor{cfrowgray}\underline{79.27} & \cellcolor{cfrowgray}\textbf{7.97} & \cellcolor{cfrowgray}\textbf{625.51} \\
\cmidrule(lr){2-15}
 & \multirow{3}{*}{$\checkmark$}
 & DMax & 92.10 & 5.48 & 557.00 & 75.40 & 5.94 & 507.00 & \underline{79.20} & 5.86 & 482.00 & 83.50 & \underline{7.36} & 637.00 \\
 & & LLaDA2.1-mini & \textbf{93.56} & \underline{5.94} & \underline{666.71} & \textbf{85.00} & \underline{7.44} & \underline{741.42} & 77.75 & \underline{7.25} & \underline{636.16} & \underline{85.98} & 7.11 & \underline{705.47} \\
 & & \cellcolor{cfrowgray}LLaDA2.1-mini-CForce (Ours) & \cellcolor{cfrowgray}\underline{92.27} & \cellcolor{cfrowgray}\textbf{7.63} & \cellcolor{cfrowgray}\textbf{845.03} & \cellcolor{cfrowgray}\underline{84.80} & \cellcolor{cfrowgray}\textbf{10.15} & \cellcolor{cfrowgray}\textbf{966.35} & \cellcolor{cfrowgray}\textbf{81.97} & \cellcolor{cfrowgray}\textbf{10.07} & \cellcolor{cfrowgray}\textbf{1016.14} & \cellcolor{cfrowgray}\textbf{86.59} & \cellcolor{cfrowgray}\textbf{8.48} & \cellcolor{cfrowgray}\textbf{870.90} \\
\bottomrule
\end{tabular}%
}
\caption{\textbf{Main Comparison by Diffusion Type and Edit Capability.} Each benchmark group reports Score, TPF, and AUP. Bold and underline mark the best and second-best metrics within each comparable block; stars mark the best pure-diffusion results.}
\label{tab:main-results}
\end{table*}

\section{Experiments}
\label{sec:experiments}

\subsection{Experimental Setup}
\label{sec:setup}

\paragraph{Models.}
We evaluate CForce on two LLaDA variants: LLaDA2.0-mini for the non-edit setting and LLaDA2.1-mini for the edit-capable setting described in Section~\ref{sec:edit}.
Detailed implementation settings are provided in Appendix~\ref{app:implementation-details}.

\paragraph{Training Data.}
We use queries from OpenMath-Instruct-2~\citep{toshniwal2025openmathinstruct} and OpenCodeInstruct~\citep{ahmad2025opencodeinstruct} and decode responses using Algorithm~\ref{alg:data}.
For LLaDA2.0, we use a confidence threshold $\tau = 0.95$; for LLaDA2.1, we use $\tau = 0.85$ and $\tau_{\mathrm{edit}} = 0.5$.
Responses exceeding 4096 tokens are discarded.

\paragraph{Evaluation.}
We evaluate on four benchmarks that span mathematical reasoning and code generation: GSM8K~\citep{cobbe2021training}, MATH500~\citep{lightman2024let}, MBPP~\citep{austin2021program}, and HumanEval~\citep{chen2021evaluating}.
Inference is performed with SGLang~\citep{zheng2024sglang} using a block size of 32 and a max generation length of 4096.
We report Score, TPF, and AUP~\citep{qian2026d3llm}. Score denotes exact-match accuracy on GSM8K and MATH500 and pass@1 on MBPP and HumanEval; higher TPF and AUP indicate greater parallelism and a better speed-quality operating point, respectively.
Appendix~\ref{app:aup-details} provides the exact operating points used to compute AUP for the LLaDA-family rows.
Throughput results on GSM8K and HumanEval, together with the hardware setup, are reported in Appendix~\ref{app:throughput}.

\paragraph{Baselines.}
We include LLaDA2.0-mini and LLaDA2.1-mini as the base models for the non-edit and edit-capable settings, respectively~\citep{bie2025llada2,bie2026llada2}.
For the LLaDA2.0-series comparison, we also include LLaDA2.0-mini-CAP, a confidence-aware parallel training variant of LLaDA2.0-mini, and refer to it as CAP~\citep{dparallel2025}.
Beyond these controlled baselines, Table~\ref{tab:main-results} reports representative recent dLLM acceleration methods: dUltra-coding-b32, d3LLM-LLaDA, d3LLM-Dream, D2F-LLaDA, LightningRL-8B-b32, and DMax~\citep{chen2025dultra,qian2026d3llm,wang2025diffusion,hu2026lightningrl,chen2026dmax}.
For LLaDA2.0-mini, CAP, and our LLaDA2.0-mini-CForce, evaluation uses $\tau = 0.9$; for LLaDA2.1-mini and LLaDA2.1-mini-CForce, evaluation uses $\tau = 0.7$ and $\tau_{\mathrm{edit}} = 0.5$.

\subsection{Main Results}
\label{sec:main-results}

\paragraph{Edit-capable dLLMs.}
Table~\ref{tab:main-results} reports controlled comparisons between each LLaDA baseline and its Consistency Forcing variant under the corresponding decoding setting, with the strongest gains appearing in the edit-capable case. On LLaDA2.1-mini, Consistency Forcing increases average TPF from 6.94 to 9.08 while also improving average score from 85.57 to 86.41.
The improvement is strongest on the code benchmarks and comes with only minor changes on the math benchmarks.
This result is consistent with the method design: edit-capable trajectories allow later corrections on already decoded tokens to feed back into the masked-position drafting ability of earlier states through the stop-gradient target.

\paragraph{Non-edit dLLMs.}
Under the non-edit decoding setting, Consistency Forcing on LLaDA2.0-mini moves the model to a higher-parallelism operating point: it reaches 6.42 average TPF, compared with 3.60 for the base model and 5.07 for CAP.
Its average score is lower than the base model but higher than CAP, making this setting an explicit speed-quality trade-off.

% \subsection{Few-Step Generation}
% \label{sec:few-step}

% \begin{table}[ht]
% \centering
% \small
% \setlength{\tabcolsep}{3pt}
% \begin{tabular}{@{}lccccc@{}}
% \toprule
% \textbf{Model} & \textbf{AVG} & \textbf{GSM8K} & \textbf{MATH} & \textbf{MBPP} & \textbf{HEval} \\
% \midrule
% \multicolumn{6}{c}{\textit{TPF = 8}} \\
% \midrule
% Vanilla & 17.13 & 35.41 & 18.80 & 8.20 & 6.10 \\
% CAP & 26.43 & 46.25 & 29.00 & 18.27 & 12.20 \\
% Ours & \textbf{38.54} & \textbf{61.33} & \textbf{54.95} & \textbf{22.01} & \textbf{15.85} \\
% \midrule
% \multicolumn{6}{c}{\textit{TPF = 4}} \\
% \midrule
% Vanilla & 54.46 & 82.79 & 70.85 & 39.81 & 24.39 \\
% CAP & 62.19 & 86.43 & 74.65 & 52.93 & 34.76 \\
% Ours & \textbf{67.19} & \textbf{88.70} & \textbf{78.10} & \textbf{53.16} & \textbf{48.78} \\
% \bottomrule
% \end{tabular}
% \caption{Few-step generation results (accuracy \%) on LLaDA2.0 series with fixed TPF budgets. Vanilla, CAP, and Ours represent LLaDA2.0-mini, LLaDA2.0-mini-CAP, and LLaDA2.0-mini-CForce respectively.}
% \label{tab:few-step}
% \end{table}

\begin{figure}[t]
\centering
\includegraphics[width=\columnwidth]{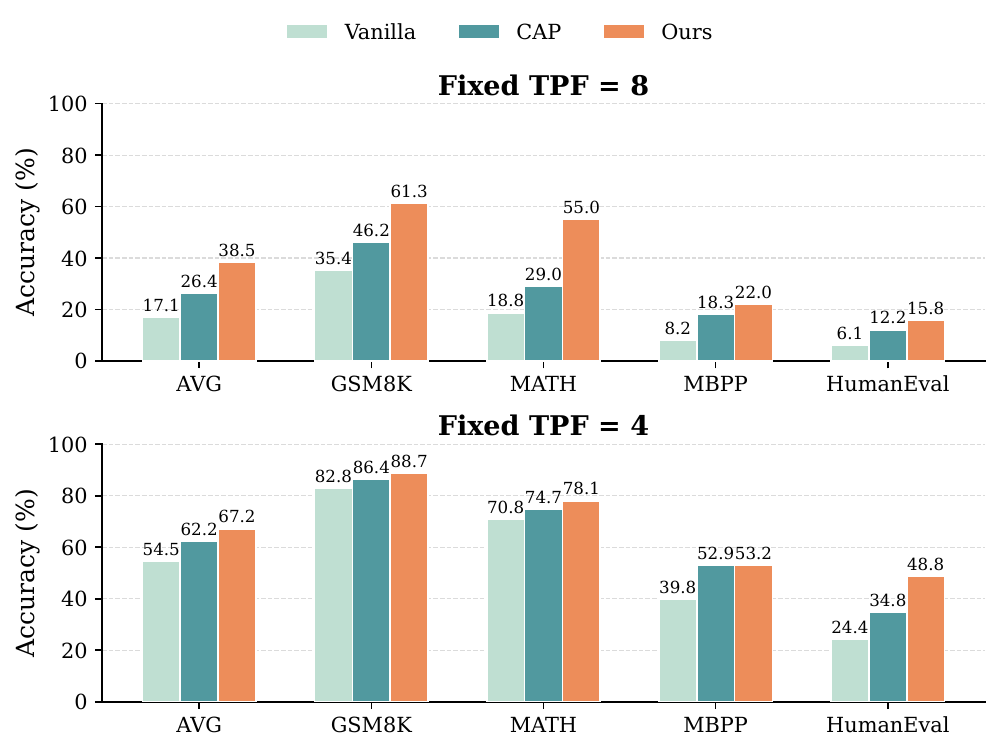}
\caption{\textbf{Few-step Generation Results (Score \%) on LLaDA2.0 Series with Fixed TPF Budgets.} Vanilla, CAP, and Ours represent LLaDA2.0-mini, LLaDA2.0-mini-CAP, and LLaDA2.0-mini-CForce respectively.}
\label{fig:few-step}
\end{figure}

\begin{table}[t]
\centering
\small
\setlength{\tabcolsep}{9pt}
\begin{tabular}{@{}ccc@{}}
\toprule
\textbf{Stage size $S$} & \textbf{AVG Score} & \textbf{AVG TPF} \\
\midrule
4 & 80.62 & 5.64 \\
\rowcolor{cfrowgray}
8 & \textbf{80.88} & 6.42 \\
16 & 75.05 & \textbf{7.97} \\
\bottomrule
\end{tabular}
\caption{\textbf{Ablation on Stage Size for LLaDA2.0-mini-CForce.} Full benchmark-level results are in Appendix~\ref{app:ablation-details}. The shaded row denotes the default setting; bold marks the best value in each metric.}
\label{tab:ablation-stage-size}
\vspace{4pt}
\centering
\small
\setlength{\tabcolsep}{3pt}
\begin{tabular}{@{}
  >{\centering\arraybackslash}p{0.21\linewidth}
  >{\centering\arraybackslash}p{0.21\linewidth}
  *{2}{>{\centering\arraybackslash}p{0.19\linewidth}}@{}}
\toprule
\multicolumn{2}{c}{\textbf{Training}} & \multicolumn{2}{c}{\textbf{AVG}} \\
\cmidrule(lr){1-2}\cmidrule(l){3-4}
\textbf{Curriculum} & \makecell{\textbf{Frozen}\\\textbf{Teacher}} & \textbf{Score} & \textbf{TPF} \\
\midrule
\rowcolor{cfrowgray}
$\checkmark$ & & \textbf{80.88} & \textbf{6.42} \\
$\checkmark$ & $\checkmark$ & 79.49 & 5.91 \\
& & 79.87 & 5.97 \\
\bottomrule
\end{tabular}
\caption{\textbf{Ablation on Curriculum Learning and Target Type for LLaDA2.0-mini-CForce.} The frozen-teacher variant changes only the target predictor; the no-curriculum variant removes the gradual exposure schedule from the same-model stop-gradient target.}
\label{tab:ablation-training}
\end{table}

% Table~\ref{tab:few-step}
\paragraph{Few-step Generation.}
\label{sec:few-step}
Figure~\ref{fig:few-step} evaluates fixed TPF decoding on the LLaDA2.0 series. We constrain the decoding budget so that the model operates at a fixed TPF of 8 or 4, which forces substantially more tokens to be committed per forward pass than in the standard dynamic-threshold setting.
Consistency Forcing obtains the best average score at both budgets, with the largest margin under the more aggressive TPF $=8$ setting.
This supports the main hypothesis that adjacent-stage forcing is most useful when early predictions must directly support large parallel commitments.

\subsection{Ablation Studies}
\label{sec:ablation}

We conduct ablations on LLaDA2.0-mini-CForce, where the confidence-threshold results expose a clear speed-quality trade-off.
We vary the stage size, curriculum schedule, KL divergence form, CE anchor weight, and target type while keeping the rest of the training setup fixed.
All rows are evaluated with threshold decoding at $\tau=0.9$ and maximum generation length 4096.

\paragraph{Stage Size.}
Table~\ref{tab:ablation-stage-size} studies the number of newly revealed tokens between saved trajectory stages.
A small stage size ($S=4$) yields nearby state pairs and lower parallelism, reaching 80.62 average score and 5.64 average TPF.
Increasing the stage size to 16 raises average TPF to 7.97 but lowers average score to 75.05, consistent with larger transitions being harder consistency targets.
We therefore use $S=8$, which provides the best observed balance in this ablation, with 80.88 average score and 6.42 average TPF.

% --- Original curriculum ablation table ---
\iffalse
\begin{table}[ht]
\centering
\small
\setlength{\tabcolsep}{3pt}
\begin{tabular}{@{}llccccc@{}}
\toprule
\textbf{Metric} & \textbf{AVG} & \textbf{GSM8K} & \textbf{MATH} & \textbf{MBPP} & \textbf{HEval} \\
\midrule
\multicolumn{6}{c}{\textit{w/ Curriculum}} \\
\midrule
Score & \textbf{80.88} & \textbf{91.74} & \textbf{79.20} & \textbf{73.30} & \textbf{79.27} \\
TPF & \textbf{6.42} & \textbf{5.36} & \textbf{6.35} & \textbf{6.00} & 7.97 \\
\midrule
\multicolumn{6}{c}{\textit{w/o Curriculum}} \\
\midrule
Score & 79.87 & 91.66 & 77.40 & 72.37 & 78.05 \\
TPF & 5.97 & 4.89 & 5.67 & 5.30 & \textbf{8.00} \\
\bottomrule
\end{tabular}
\caption{Ablation on curriculum learning. Curriculum training improves both accuracy (+1.01 avg) and TPF (+0.45 avg).}
\label{tab:ablation-curriculum}
\end{table}
\fi
% --- End of original curriculum ablation table ---

\paragraph{Curriculum Learning.}
Table~\ref{tab:ablation-training} isolates the curriculum schedule (Eq.~\ref{eq:curriculum}), which gradually increases the fraction of newly revealed positions exposed by the later trajectory state from 10\% to 100\% over training.
Removing the schedule lowers average score from 80.88 to 79.87 and average TPF from 6.42 to 5.97, confirming that gradual exposure to larger context gaps stabilizes learning.
This suggests that the model benefits from first learning local adjacent-stage consistency before being exposed to the full transition gap.

\paragraph{Target Type.}
Table~\ref{tab:ablation-training} also compares the same-model stop-gradient target used by Consistency Forcing with a frozen-teacher variant.
The same-model target outperforms the frozen-teacher baseline on both score (+1.39) and TPF (+0.51), suggesting that an evolving target better tracks the student's own decoding distribution and provides more aligned supervision for its inference states.

\begin{table}[ht]
\centering
\small
\setlength{\tabcolsep}{6pt}
\begin{tabular}{@{}llcc@{}}
\toprule
\multirow{2}{*}{\textbf{Component}} & \multirow{2}{*}{\textbf{Variant}} & \multicolumn{2}{c}{\textbf{AVG}} \\
\cmidrule(l){3-4}
& & \textbf{Score} & \textbf{TPF} \\
\midrule
\multirow{3}{*}{\makecell[l]{KL\\divergence}}
 & Forward KL & 85.75 & 5.07 \\
 & Reverse KL & 65.57 & 8.72 \\
 & \cellcolor{cfrowgray}CAD & \cellcolor{cfrowgray}80.88 & \cellcolor{cfrowgray}6.42 \\
\midrule
\multirow{3}{*}{\makecell[l]{CE\\anchor}}
 & $\lambda_{\mathrm{CE}}=0$ & 79.02 & 6.52 \\
 & \cellcolor{cfrowgray}$\lambda_{\mathrm{CE}}=0.1$ & \cellcolor{cfrowgray}80.88 & \cellcolor{cfrowgray}6.42 \\
 & $\lambda_{\mathrm{CE}}=2.0$ & 80.22 & 6.52 \\
\bottomrule
\end{tabular}
\caption{\textbf{Ablations of the KL Divergence Form and CE Anchor Weight on LLaDA2.0-mini-CForce.}}
\label{tab:ablation-objective}
\end{table}

% --- Original objective ablation table ---
\iffalse
\begin{table*}[t]
\centering
\small
\setlength{\tabcolsep}{4pt}
\begin{tabular}{@{}lllccccc@{}}
\toprule
\textbf{Component} & \textbf{Variant} & \textbf{Metric} & \textbf{AVG} & \textbf{GSM8K} & \textbf{MATH} & \textbf{MBPP} & \textbf{HEval} \\
\midrule
\multirow{6}{*}{KL divergence}
 & \multirow{2}{*}{Forward KL} & Score & 85.75 & 92.95 & 82.80 & 81.26 & 85.98 \\
 &  & TPF & 5.07 & 4.00 & 4.72 & 4.74 & 6.81 \\
 & \multirow{2}{*}{Reverse KL} & Score & 65.57 & 87.64 & 70.20 & 51.99 & 52.44 \\
 &  & TPF & 8.72 & 7.02 & 8.15 & 8.78 & 10.92 \\
 & \multirow{2}{*}{CAD} & Score & 80.88 & 91.74 & 79.20 & 73.30 & 79.27 \\
 &  & TPF & 6.42 & 5.36 & 6.35 & 6.00 & 7.97 \\
\midrule
\multirow{6}{*}{CE anchor}
 & \multirow{2}{*}{$\lambda_{\mathrm{CE}}=0$} & Score & 79.02 & 91.58 & 78.80 & 68.85 & 76.83 \\
 &  & TPF & 6.52 & 5.34 & 6.36 & 6.06 & 8.32 \\
 & \multirow{2}{*}{$\lambda_{\mathrm{CE}}=0.1$} & Score & 80.88 & 91.74 & 79.20 & 73.30 & 79.27 \\
 &  & TPF & 6.42 & 5.36 & 6.35 & 6.00 & 7.97 \\
 & \multirow{2}{*}{$\lambda_{\mathrm{CE}}=2.0$} & Score & 80.22 & 92.49 & 78.60 & 75.41 & 74.39 \\
 &  & TPF & 6.52 & 5.34 & 6.06 & 6.34 & 8.35 \\
\bottomrule
\end{tabular}
\caption{Ablations of the KL divergence form and CE anchor weight on LLaDA2.0-mini-CForce. These results show how each objective choice shifts the score--TPF operating point.}
\label{tab:ablation-objective}
\end{table*}
\fi
% --- End of original objective ablation table ---

\paragraph{KL Divergence Form.}
Table~\ref{tab:ablation-objective} compares forward KL, reverse KL, and our CAD.
Forward KL gives the highest average score, but it remains conservative in parallelism, with 5.07 average TPF.
Reverse KL reaches much higher parallelism, 8.72 average TPF, but its average score drops to 65.57, consistent with an overly aggressive mode-seeking objective under threshold decoding.
CAD selects an intermediate operating point: compared with forward KL, it increases average TPF from 5.07 to 6.42; compared with reverse KL, it avoids the large score collapse.
CAD thus balances conservative drift control with controlled sharpening.

\paragraph{CE Anchor.}
The lower block of Table~\ref{tab:ablation-objective} studies the CE anchor weight.
Removing the anchor lowers average score from 80.88 to 79.02 while leaving TPF nearly unchanged, suggesting that the CE term improves the score side of the trade-off.
A larger weight, $\lambda_{\mathrm{CE}}=2.0$, recovers part of the score but remains below the moderate setting on average and drops noticeably on HumanEval.
We therefore use $\lambda_{\mathrm{CE}}=0.1$ as a modest token-level anchor: it improves average score while preserving the adjacent-stage KL-driven parallelism.

%% file: sections/conclusion.tex
\section{Conclusion}
\label{sec:conclusion}

We presented Consistency Forcing, a training method for improving the speed-quality frontier of diffusion language models under aggressive threshold decoding. The method constructs staged trajectories from the model's own decoding path and aligns earlier low-context states with later, more informative states through same-model stop-gradient supervision. By combining Confidence Adaptive KL Divergence, a CE anchor, and a curriculum over reveal difficulty, Consistency Forcing improves the reliability of early-stage predictions and supports higher parallelism. We further extended the framework to edit-capable dLLMs, where later T2T refinements provide useful supervision for earlier M2T predictions. Experiments on both edit-capable and conventional dLLMs show that this trajectory-based forcing improves high-parallelism decoding while preserving the generation quality.

%% file: sections/limitations.tex
\section*{Limitations}

For training stability, Consistency Forcing currently relies on trajectories collected in advance from a pretrained initialization. Although these trajectories are produced by the model's own threshold-decoding process, they are still fixed before optimization begins. As training changes the model, the cached trajectories may gradually differ from the actual inference trajectories visited by the updated model. This remaining mismatch can limit how closely the training signal follows the final student distribution. In future work, we plan to explore online trajectory collection and policy-updating strategies so that the forcing objective can adapt to the model's evolving inference behavior.

%% file: sections/acknowledgments.tex
\section*{Acknowledgments}

This work was supported by Ant Group Research Fund.

%% file: sections/apdx.tex
\section{Staged Trajectory Construction Algorithm}
\label{app:trajectory-algorithm}

Algorithm~\ref{alg:data} gives the complete procedure for constructing staged self-rollout trajectories from native threshold-decoding traces.

\begin{algorithm}[H]
\caption{Staged Trajectory Construction}\label{alg:data}
\begin{algorithmic}
\small
\Require Pretrained dLLM $q_{\theta_0}$, prompt set $\mathcal{X}$, M2T threshold $\tau$, optional edit threshold $\tau_{\mathrm{edit}}$, stage size $S$
\Ensure Trajectory dataset $\mathcal{D}$
\Statex \textcolor{cforange}{// \textbf{Initialize dataset}}
\State $\mathcal{D}\gets\emptyset$
\For{each prompt $x\in\mathcal{X}$}
\Statex \hspace{\algorithmicindent}\textcolor{cforange}{// \textbf{Initialize native trajectory}}
\State Set $s_0$ to the all-mask initial state and save $\bar{s}_0\gets s_0$
\State Set stage counter $n\gets0$ and reveal counter $R\gets0$
\For{each native decoding step $t$}
    \State Compute predictions with $q_{\theta_0}(\cdot\mid x,s_t,j)$
    \State Select newly revealed positions $\Delta M_t$ by Eq.~\eqref{eq:threshold}
    \State For edit-capable decoding, record $E_t$ by Eq.~\eqref{eq:edit}
    \State Update the native state from $s_t$ to $s_{t+1}$
    \State Set $R\gets R+|\Delta M_t|$
    \If{$R$ reaches the stage size $S$}
        \State Save $\bar{s}_{n+1}\gets s_{t+1}$
        \State Set $n\gets n+1$ and $R\gets0$
    \EndIf
\EndFor
\Statex \hspace{\algorithmicindent}\textcolor{cforange}{// \textbf{Finalize after full decoding}}
\If{$s_T$ is not already saved}
    \State Save $\bar{s}_N\gets s_T$
\EndIf
\State Add the complete stage trajectory $\bar{\mathcal{T}}=(\bar{s}_0,\ldots,\bar{s}_N)$ to $\mathcal{D}$
\EndFor
\State \Return $\mathcal{D}$
\end{algorithmic}
\end{algorithm}

\section{Implementation Details}
\label{app:implementation-details}

All models use full-parameter fine-tuning with the AdamW optimizer, a learning rate of $1.0 \times 10^{-5}$, and a cosine learning rate schedule.
We set the stage size to 8, use a global batch size of 64, and train for 5 epochs.
The CE anchor loss weight $\lambda_{\mathrm{CE}}$ is set to 0.1, and the block size is 32.
During KL distillation, we retain the top-20 tokens ranked by the target distribution $p_{\mathrm{tar}}^j$ and renormalize the retained probabilities before computing the KL terms.
This top-$k$ truncation both reduces the computation cost of the softmax and KL terms and focuses the student on the semantically meaningful region of the distribution.
All training is conducted on 64 NVIDIA H800 GPUs.
% TODO: Add GPU-hours once wall-clock training logs are available.

\section{Full Derivation of the Theoretical Analysis}
\label{app:theory}

This appendix provides the full derivation of the bound used in Section~\ref{sec:theory}. Using the TV distance defined in Section~\ref{sec:theory}, consider a position $j$ that is revealed at stage $r_j$, i.e., $j\in \bar{M}_{r_j-1}$ and $j\notin \bar{M}_{r_j}$, with final token $y^j=\bar{s}_N^j$. Let $p_i^j=q_\theta(\cdot\mid x,\bar{s}_i,j)$ denote the prediction at stage $i$. For any earlier masked stage $i<r_j$, the triangle inequality gives
\begin{equation}\label{eq:app-tv-decomp}
\small
D_{\mathrm{TV}}(p_i^j,\delta_{y^j})
\le
\sum_{r=i}^{r_j-2}D_{\mathrm{TV}}(p_r^j,p_{r+1}^j)
+D_{\mathrm{TV}}(p_{r_j-1}^j,\delta_{y^j}),
\end{equation}
where $\delta_{y^j}$ is the one-hot distribution on $y^j$. The first term accumulates distributional drift across adjacent stages, and the second term is the masked-state error immediately before token reveal.

Pinsker's inequality bounds each adjacent-stage TV term by forward KL:
\begin{equation}\label{eq:app-pinsker}
D_{\mathrm{TV}}(p_r^j,p_{r+1}^j)
\le
\sqrt{\frac{1}{2}D_{\mathrm{KL}}(p_{r+1}^j\|p_r^j)}.
\end{equation}
For the reveal-boundary term, we have
\begin{equation}\label{eq:app-ce-bound}
\begin{split}
D_{\mathrm{TV}}(p_{r_j-1}^j,\delta_{y^j})
&= 1-p_{r_j-1}^j(y^j) \\
&\le -\log p_{r_j-1}^j(y^j),
\end{split}
\end{equation}
using $1-x\le-\log x$ for $x\in(0,1]$. Substituting Eqs.~\eqref{eq:app-pinsker} and~\eqref{eq:app-ce-bound} into Eq.~\eqref{eq:app-tv-decomp} yields
\begin{equation}\label{eq:app-combined-bound}
\small
\begin{split}
D_{\mathrm{TV}}(p_i^j,\delta_{y^j})
&\le
\sum_{r=i}^{r_j-2}
\sqrt{\frac{1}{2}D_{\mathrm{KL}}(p_{r+1}^j\|p_r^j)} \\
&\quad - \log p_{r_j-1}^j(y^j).
\end{split}
\end{equation}

This bound matches the structure of Consistency Forcing. The forward KL component in Eq.~\eqref{eq:cad} directly targets the adjacent-stage drift terms in Eq.~\eqref{eq:app-combined-bound}. The CE anchor in Eq.~\eqref{eq:ce} includes the reveal-boundary term in Eq.~\eqref{eq:app-ce-bound} and supplies token-level supervision at commitment boundaries. The confidence-weighted reverse KL further encourages mode-seeking predictions when the later-stage target is reliable. Since $c_j\ge0$,
\begin{equation}
\small
\begin{split}
\mathcal{L}_{\mathrm{CAD}}^j
&= D_{\mathrm{KL}}(p_{\mathrm{tar}}^j\|p_{\mathrm{stu}}^j)
+ c_jD_{\mathrm{KL}}(p_{\mathrm{stu}}^j\|p_{\mathrm{tar}}^j) \\
&\ge D_{\mathrm{KL}}(p_{\mathrm{tar}}^j\|p_{\mathrm{stu}}^j),
\end{split}
\end{equation}
so optimizing $\mathcal{L}_{\mathrm{CAD}}$ preserves the forward KL alignment term while sharpening the student distribution toward the cleaner state's dominant mode. Therefore, Consistency Forcing reduces a tractable surrogate of the early-stage prediction error bound by aligning adjacent-stage distributions and anchoring masked-token predictions along the trajectory.

\section{AUP Evaluation Details}
\label{app:aup-details}

\begin{table*}[ht!]
\centering
\scriptsize
\setlength{\tabcolsep}{2.4pt}
\resizebox{\textwidth}{!}{%
\begin{tabular}{@{}ll*{5}{cc}@{}}
\toprule
\multirow{2}{*}{\textbf{Model}} & \multirow{2}{*}{\textbf{Decode config}}
& \multicolumn{2}{c}{\textbf{AVG}}
& \multicolumn{2}{c}{\textbf{GSM8K}}
& \multicolumn{2}{c}{\textbf{MATH500}}
& \multicolumn{2}{c}{\textbf{MBPP}}
& \multicolumn{2}{c}{\textbf{HumanEval}} \\
\cmidrule(lr){3-4}\cmidrule(lr){5-6}\cmidrule(lr){7-8}\cmidrule(lr){9-10}\cmidrule(l){11-12}
& & \textbf{Score} & \textbf{TPF}
& \textbf{Score} & \textbf{TPF}
& \textbf{Score} & \textbf{TPF}
& \textbf{Score} & \textbf{TPF}
& \textbf{Score} & \textbf{TPF} \\
\midrule
\multirow{2}{*}{LLaDA2.0-mini} & $\tau=0.95$
& \textbf{86.52} & 3.09 & \textbf{93.78} & 2.42 & \textbf{84.20} & 2.99 & \textbf{81.50} & 3.21 & \textbf{86.59} & 3.74 \\
& $\tau=0.9$
& \underline{84.32} & 3.60 & \underline{93.25} & 2.85 & 81.80 & 3.46 & 78.69 & 3.80 & \underline{83.54} & 4.28 \\
\multirow{2}{*}{LLaDA2.0-mini-CAP} & $\tau=0.95$
& 82.70 & 4.26 & 92.34 & 3.40 & \underline{83.00} & 4.15 & 76.81 & 4.57 & 78.66 & 4.93 \\
& $\tau=0.9$
& 79.51 & 5.07 & 91.74 & 4.09 & 81.00 & 4.85 & 72.13 & \underline{5.43} & 73.18 & 5.89 \\
\multirow{2}{*}{LLaDA2.0-mini-CForce (Ours)} & $\tau=0.95$
& 83.90 & \underline{5.32} & 92.27 & \underline{4.33} & 82.60 & \underline{5.34} & \underline{79.63} & 4.87 & 81.10 & \underline{6.73} \\
& $\tau=0.9$
& 80.88 & \textbf{6.42} & 91.74 & \textbf{5.36} & 79.20 & \textbf{6.35} & 73.30 & \textbf{6.00} & 79.27 & \textbf{7.97} \\
\midrule
\multirow{2}{*}{LLaDA2.1-mini} & $\tau=0.7$, $\tau_{\mathrm{edit}}=0.5$
& 85.57 & 6.94 & \textbf{93.56} & 5.94 & \underline{85.00} & 7.44 & 77.75 & 7.25 & \underline{85.98} & 7.11 \\
& $\tau=0.5$, $\tau_{\mathrm{edit}}=0.0$
& \underline{85.77} & 8.11 & \underline{92.95} & 7.15 & \textbf{86.20} & 8.74 & \underline{79.16} & 8.30 & 84.76 & 8.26 \\
\multirow{2}{*}{LLaDA2.1-mini-CForce (Ours)} & $\tau=0.7$, $\tau_{\mathrm{edit}}=0.5$
& \textbf{86.41} & \underline{9.08} & 92.27 & \underline{7.63} & 84.80 & \underline{10.15} & \textbf{81.97} & \underline{10.07} & \textbf{86.59} & \underline{8.48} \\
& $\tau=0.5$, $\tau_{\mathrm{edit}}=0.0$
& 85.03 & \textbf{10.86} & 91.51 & \textbf{9.26} & 84.40 & \textbf{11.47} & 78.22 & \textbf{12.61} & \underline{85.98} & \textbf{10.08} \\
\bottomrule
\end{tabular}%
}
\caption{\textbf{Benchmark-level Operating Points Used to Compute AUP for the LLaDA-family Rows in Table~\ref{tab:main-results}.} Each benchmark group reports the Score/TPF pair used by the two-point AUP frontier. Bold and underline mark the best and second-best metrics within each LLaDA-series block. All evaluations use maximum generation length 4096; LLaDA2.0 models are M2T-only and therefore do not use an edit threshold.}
\label{tab:aup-frontier-points}
\end{table*}

AUP is computed from a small score--parallelism frontier rather than from only the single Score/TPF pair shown in Table~\ref{tab:main-results}. For each benchmark, an operating point is denoted by $(a_i,y_i)$, where $a_i$ is TPF and $y_i$ is the task score in percentage points. Following the public AUP implementation, we sort operating points by TPF and compute
\begin{equation}
\small
\mathrm{AUP}
= a_1 y_1
+ \sum_{i=2}^{m}
\frac{a_i-a_{i-1}}{2}
\left(y_i W(y_i)+y_{i-1}W(y_{i-1})\right),
\end{equation}
where $W(y)=\min(\exp(-3(1-y/y_{\max})),1)$ and $y_{\max}$ is the maximum score for that benchmark among the compared methods. We keep operating points within five score points of the first, lowest-TPF point, matching the default AUP filtering threshold. This weighting penalizes operating points that gain parallelism by substantially degrading accuracy.

Table~\ref{tab:aup-frontier-points} lists the exact decoding configurations and benchmark-level Score/TPF pairs used to construct the LLaDA-family AUP frontiers in Table~\ref{tab:main-results}. For the LLaDA2.0 series, including LLaDA2.0-mini, LLaDA2.0-mini-CAP, and LLaDA2.0-mini-CForce, we evaluate two threshold-decoding points, $\tau=0.95$ and $\tau=0.9$. For the edit-capable LLaDA2.1 series, including LLaDA2.1-mini and LLaDA2.1-mini-CForce, we use the official quality and speed modes: quality mode sets $\tau=0.7$ and $\tau_{\mathrm{edit}}=0.5$, while speed mode sets $\tau=0.5$ and $\tau_{\mathrm{edit}}=0.0$. The Score and TPF columns in Table~\ref{tab:main-results} report the main comparison operating point, while the AUP column is computed from the corresponding two-point frontier.

\section{Wall-clock Throughput}
\label{app:throughput}

We report wall-clock throughput on two benchmarks, GSM8K and HumanEval, in Table~\ref{tab:throughput}. All measurements use two NVIDIA H20 GPUs with tensor parallelism $\mathrm{TP}=2$ and batch size 1. TPS denotes generated tokens per second.

\begin{table}[t]
\centering
\small
\setlength{\tabcolsep}{4pt}
\resizebox{\columnwidth}{!}{%
\begin{tabular}{@{}lccc@{}}
\toprule
\textbf{Model} & \textbf{GSM8K TPS} & \textbf{HumanEval TPS} & \textbf{AVG TPS} \\
\midrule
\multicolumn{4}{c}{\textit{Non-edit dLLMs}} \\
\midrule
LLaDA2.0-mini & 489.87 & 711.59 & 600.73 \\
LLaDA2.0-mini-CAP & 710.88 & 891.94 & 801.41 \\
LLaDA2.0-mini-CForce (Ours) & \textbf{783.09} & \textbf{932.82} & \textbf{857.96} \\
\midrule
\multicolumn{4}{c}{\textit{Edit-capable dLLMs}} \\
\midrule
LLaDA2.1-mini & 987.78 & 1116.10 & 1051.94 \\
LLaDA2.1-mini-CForce (Ours) & \textbf{1231.52} & \textbf{1352.86} & \textbf{1292.19} \\
\bottomrule
\end{tabular}%
}
\caption{\textbf{Wall-clock Throughput on GSM8K and HumanEval.} Bold marks the best throughput within each edit-capability group.}
\label{tab:throughput}
\end{table}

In the non-edit LLaDA2.0 setting, CForce reaches 857.96 average TPS, improving over the base model by 42.82\% and over CAP by 7.06\%. In the edit-capable LLaDA2.1 setting, CForce reaches 1292.19 average TPS, improving over the base model by 22.84\%.

\section{Full Ablation Results}
\label{app:ablation-details}

Tables~\ref{tab:full-ablation-stage-size}, \ref{tab:full-ablation-training}, and~\ref{tab:full-ablation-objective} provide the benchmark-level results corresponding to the compact ablation tables in Section~\ref{sec:ablation}.
Each benchmark group reports Score and TPF under threshold decoding at $\tau=0.9$ and maximum generation length 4096.

\begin{table*}[ht!]
\centering
\fontsize{5}{5.5}\selectfont
\setlength{\tabcolsep}{2.4pt}
\resizebox{\textwidth}{!}{%
\begin{tabular}{@{}c*{5}{cc}@{}}
\toprule
\multirow{2}{*}{\textbf{Stage size $S$}}
& \multicolumn{2}{c}{\textbf{AVG}}
& \multicolumn{2}{c}{\textbf{GSM8K}}
& \multicolumn{2}{c}{\textbf{MATH500}}
& \multicolumn{2}{c}{\textbf{MBPP}}
& \multicolumn{2}{c}{\textbf{HumanEval}} \\
\cmidrule(lr){2-3}\cmidrule(lr){4-5}\cmidrule(lr){6-7}\cmidrule(lr){8-9}\cmidrule(l){10-11}
& \textbf{Score} & \textbf{TPF}
& \textbf{Score} & \textbf{TPF}
& \textbf{Score} & \textbf{TPF}
& \textbf{Score} & \textbf{TPF}
& \textbf{Score} & \textbf{TPF} \\
\midrule
4 & 80.62 & 5.64 & 91.51 & 4.56 & 79.80 & 5.64 & 74.94 & 5.22 & 76.22 & 7.14 \\
\rowcolor{cfrowgray}
8 & 80.88 & 6.42 & 91.74 & 5.36 & 79.20 & 6.35 & 73.30 & 6.00 & 79.27 & 7.97 \\
16 & 75.05 & 7.97 & 90.30 & 7.61 & 75.20 & 9.86 & 62.76 & 6.71 & 71.95 & 7.70 \\
\bottomrule
\end{tabular}%
}
\caption{\textbf{Full benchmark-level ablation on stage size for LLaDA2.0-mini-CForce.} Each benchmark group reports Score and TPF.}
\label{tab:full-ablation-stage-size}
\end{table*}

\begin{table*}[ht!]
\centering
\scriptsize
\setlength{\tabcolsep}{2.4pt}
\resizebox{\textwidth}{!}{%
\begin{tabular}{@{}ll*{5}{cc}@{}}
\toprule
\multirow{2}{*}{\textbf{Curriculum}} & \multirow{2}{*}{\textbf{Target type}}
& \multicolumn{2}{c}{\textbf{AVG}}
& \multicolumn{2}{c}{\textbf{GSM8K}}
& \multicolumn{2}{c}{\textbf{MATH500}}
& \multicolumn{2}{c}{\textbf{MBPP}}
& \multicolumn{2}{c}{\textbf{HumanEval}} \\
\cmidrule(lr){3-4}\cmidrule(lr){5-6}\cmidrule(lr){7-8}\cmidrule(lr){9-10}\cmidrule(l){11-12}
& & \textbf{Score} & \textbf{TPF}
& \textbf{Score} & \textbf{TPF}
& \textbf{Score} & \textbf{TPF}
& \textbf{Score} & \textbf{TPF}
& \textbf{Score} & \textbf{TPF} \\
\midrule
\rowcolor{cfrowgray}
$\checkmark$ & Same-model target & 80.88 & 6.42 & 91.74 & 5.36 & 79.20 & 6.35 & 73.30 & 6.00 & 79.27 & 7.97 \\
$\checkmark$ & Frozen teacher & 79.49 & 5.91 & 91.21 & 4.67 & 79.20 & 5.57 & 70.73 & 5.63 & 76.83 & 7.75 \\
$\times$ & Same-model target & 79.87 & 5.97 & 91.66 & 4.89 & 77.40 & 5.67 & 72.37 & 5.30 & 78.05 & 8.00 \\
\bottomrule
\end{tabular}%
}
\caption{\textbf{Full benchmark-level ablation on curriculum learning and target type for LLaDA2.0-mini-CForce.} The frozen-teacher variant changes only the later-state target predictor, while the no-curriculum row keeps the same-model stop-gradient target and removes the gradual exposure schedule. Each benchmark group reports Score and TPF.}
\label{tab:full-ablation-training}
\end{table*}

\begin{table*}[ht!]
\centering
\scriptsize
\setlength{\tabcolsep}{2.4pt}
\resizebox{\textwidth}{!}{%
\begin{tabular}{@{}ll*{5}{cc}@{}}
\toprule
\multirow{2}{*}{\textbf{Component}} & \multirow{2}{*}{\textbf{Variant}}
& \multicolumn{2}{c}{\textbf{AVG}}
& \multicolumn{2}{c}{\textbf{GSM8K}}
& \multicolumn{2}{c}{\textbf{MATH500}}
& \multicolumn{2}{c}{\textbf{MBPP}}
& \multicolumn{2}{c}{\textbf{HumanEval}} \\
\cmidrule(lr){3-4}\cmidrule(lr){5-6}\cmidrule(lr){7-8}\cmidrule(lr){9-10}\cmidrule(l){11-12}
& & \textbf{Score} & \textbf{TPF}
& \textbf{Score} & \textbf{TPF}
& \textbf{Score} & \textbf{TPF}
& \textbf{Score} & \textbf{TPF}
& \textbf{Score} & \textbf{TPF} \\
\midrule
\multirow{3}{*}{KL divergence}
& Forward KL & 85.75 & 5.07 & 92.95 & 4.00 & 82.80 & 4.72 & 81.26 & 4.74 & 85.98 & 6.81 \\
& Reverse KL & 65.57 & 8.72 & 87.64 & 7.02 & 70.20 & 8.15 & 51.99 & 8.78 & 52.44 & 10.92 \\
& \cellcolor{cfrowgray}CAD & \cellcolor{cfrowgray}80.88 & \cellcolor{cfrowgray}6.42 & \cellcolor{cfrowgray}91.74 & \cellcolor{cfrowgray}5.36 & \cellcolor{cfrowgray}79.20 & \cellcolor{cfrowgray}6.35 & \cellcolor{cfrowgray}73.30 & \cellcolor{cfrowgray}6.00 & \cellcolor{cfrowgray}79.27 & \cellcolor{cfrowgray}7.97 \\
\midrule
\multirow{3}{*}{CE anchor}
& $\lambda_{\mathrm{CE}}=0$ & 79.02 & 6.52 & 91.58 & 5.34 & 78.80 & 6.36 & 68.85 & 6.06 & 76.83 & 8.32 \\
& \cellcolor{cfrowgray}$\lambda_{\mathrm{CE}}=0.1$ & \cellcolor{cfrowgray}80.88 & \cellcolor{cfrowgray}6.42 & \cellcolor{cfrowgray}91.74 & \cellcolor{cfrowgray}5.36 & \cellcolor{cfrowgray}79.20 & \cellcolor{cfrowgray}6.35 & \cellcolor{cfrowgray}73.30 & \cellcolor{cfrowgray}6.00 & \cellcolor{cfrowgray}79.27 & \cellcolor{cfrowgray}7.97 \\
& $\lambda_{\mathrm{CE}}=2.0$ & 80.22 & 6.52 & 92.49 & 5.34 & 78.60 & 6.06 & 75.41 & 6.34 & 74.39 & 8.35 \\
\bottomrule
\end{tabular}%
}
\caption{\textbf{Full benchmark-level ablations of the KL Divergence Form and CE Anchor Weight on LLaDA2.0-mini-CForce.} Each benchmark group reports Score and TPF.}
\label{tab:full-ablation-objective}
\end{table*}

\section{Case Study}
\label{app:case-study}

We present qualitative examples comparing the generation quality of LLaDA2.0-mini-CForce (Ours), LLaDA2.0-mini-CAP, and the LLaDA2.0-mini baseline. All outputs are generated with a fixed TPF of 8, i.e., each forward pass produces exactly 8 tokens. These examples illustrate that CForce produces more coherent and accurate outputs under few-step generation compared to CAP and the baseline.

\subsection{Example 1 (Reference: 140)}

\textbf{Prompt:}
\begin{lstlisting}
Question: Andy plants 90 geraniums and 40 fewer petunias that geraniums. How many flowers does he plant total?
Please reason step by step, and put your final answer within \boxed{}.
\end{lstlisting}

\textbf{LLaDA2.0-mini-CForce (Ours):}
\begin{lstlisting}
Let's solve the problem step by step:

1. **Number of geraniums**:
   Andy plants 90 geraniums.

2. **Number of petunias**:
   He plants 40 fewer petunias than geraniums.
   So, number of petunias = 90 - 40 = 50.

3. **Total number of flowers**:
   Add ger geranium and ger petunias:
   Total = geraniums + petunias = 90 + 50 = 140.

Final answer:
\boxed{140}
\end{lstlisting}

\textbf{LLaDA2.0-mini-CAP:}
\begin{lstlisting}
Let's break this down step by step:

1. **Identify the number of geraniums planted:**
   Andy plants **90 geraniums**.

2. **Determine the number of petunias planted:**
   He plants **40 fewer petunias than geraniums**.
   So, number of petunias = 90 geraniumaniums)0 = **50**.

3. **Calculate the total number of flowers:**
   Total flowers = geraniums + petunias
   = 900 + 50 =
   **120

\boxed{Rightarrow{1}0}
\end{lstlisting}

\textbf{LLaDA2.0-mini (Baseline):}
\begin{lstlisting}
Let's solve this problem step by step.

1. **Identify the number of geraniums plants**:
   Andy plants **90 geraniums**.

2. **Determine the number of petunias he plants**:
 He plants **40 fewer petununias than geraniums**.
   So, number of petunias =
   $$
   90 - 040 = 50
   $$

3. **Calculate the total number of flowers**:
   Add flowers number of geraniums and petunias
   $$
   90 + 50 = 140
   $$

Wait -- the total written.:
\ $$
   \boxed - 50 = 130 \\
   $$

No -- let's re-check.

**90 - 40 = 50**, correct.5
ias total.

$$
\ {90 } = \1100
$$

No -- restart.

Correct:
$$
90 + 50 = 140
$$

Correcting \boxed{140}

After Final step:

\boxed$$
140
$$
\end{lstlisting}

\subsection{Example 2 (Reference: 160)}

\textbf{Prompt:}
\begin{lstlisting}
Question: Tara bought 8 packs of 5 canvas bags for $4 each. She painted them and sold them at a craft fair for $8 each. How much profit did she earn on her bags?
Please reason step by step, and put your final answer within \boxed{}.
\end{lstlisting}

\textbf{LLaDA2.0-mini-CForce (Ours):}
\begin{lstlisting}
Let's solve this step by step.

**Step 1: Determine the total number of canvas bags Tara bought.**

Tara bought 8 packs, and each pack contains 5 canvas bags:

$$
8 \text{ packs} \times 5 \text{ bags per pack} = 40 \text{ canvas bags}
$$

**Step 2: Calculate the total cost of the canvas bags.**

Each canvas bag costs $4:

$$
40 \text{ bags} \times \$4 = \$160
$$

**Step 3: Calculate the total revenue from selling the bags.**

She sold each bag for $8:

$$
40 \text{ bags} \times \$8 = \$320
$$

**Step 4: Calculate the profit.**

Profit = Revenue - Total Cost

$$
\$320 - \$160 = \$160
$$

**Final Answer:**

$$
\boxed{160}
$$

Tara earned a profit of **$160**.
\end{lstlisting}

\textbf{LLaDA2.0-mini-CAP:}
\begin{lstlisting}
Let's solve this step step by step:

**Step 1: Calculate the total cost Tara the bags.**

 bought bought 8 packs of 5 canvas bags,
 pack. Each pack pack costs $4.

So, total number of bags bags =
8 packs x{5 bags per pack} = {4{text{ bags}
$$

**Total cost** =
8 x {44 per pack} = \{32}
$$

---

**Step 2: Calculate the much revenue selling the the bags bags
$$
She sold each 40 bags for $8 each.

Total, from all bags =
40 bags x {$8 per bag} \ \{320}
$$

---

**Step 3: Calculate the profit
$$ProfitProfit = Total Revenue - Total cost
 \{320 - \$32} = \{288}

---

Wait -- correction:

$$
$$

boxedboxed2
\end{lstlisting}

\textbf{LLaDA2.0-mini (Baseline):}
\begin{lstlisting}
Let's break this down step by step:

**Step 1: Calculate how total canvas bags Tara bought.**
T bought bought 8 packs, each of 5 canvas bags.

So,
Total \text {5}-canvas bags} = 8 x packs 5 = =text40 bags.

**Step 2: Calculate how Tara cost for bought on canvas bags.**
Each pack $44, and she bought 8..

So,
Cost = 8 x $4 = \text3$2.

**Step 3 Calculate how much Tara earned from selling the bags.**
She sold each bag bag for for for88, and had 40 bags.
.

So, totalRevenue = 40 x $8 = \text320.

**Step 4: profit earned vs.**
Profit = Total revenue - Total Cost.

Profit = $320 - 32 = \text288..

Wait -- let's correct and recal values accurately.
\end{lstlisting}

\subsection{Example 3}

\textbf{Prompt:}
\begin{lstlisting}
You are an expert Python programmer, and here is your task: Write a function to find the median of three numbers. Your code should pass these tests:

 assert median_numbers(25,55,65)==55.0
assert median_numbers(20,10,30)==20.0
assert median_numbers(15,45,75)==45.0
Please enclose your code within delimiters as follows:
```python
# YOUR CODE HERE
```
\end{lstlisting}

\textbf{LLaDA2.0-mini-CForce (Ours):}
\begin{lstlisting}
```python
def median_numbers(a, b, c):
    """
    Find the median of three numbers.

    Args:
        a, b, c: three numbers

    Returns:
        The: of a float
    """

    # Sort a three numbers list and return the median
    sorted_numbers = sorted([a, b, c])
    return float(sorted_numbers[1])
```
\end{lstlisting}

\textbf{LLaDA2.0-mini-CAP:}
\begin{lstlisting}
```python
def median_numbers(a,,, c):
    """
    Find the median of three numbers.

    Args:
        a, b, c: Three numbers

    Returns:
        The median value as a numbers or a float
    place
    numbers = [a, b, c]
    numbers.sort()
    return float(numbers[1])
```
\end{lstlisting}

\textbf{LLaDA2.0-mini (Baseline):}
\begin{lstlisting}
```python
# median_numbers(a, b, c):
    """
    Find the median of three numbers.

    Args:
        a, b, c: Three numbers

    Returns:
        The median as a three-digit float
    float

    # Create a three numbers numbers and sort
    numbers = [] + [b] +cc
    numbers.sort()

    # Return the median (middle value)
    return float(numbers[1])
```
\end{lstlisting}